%% file: document.tex
\documentclass[journal,transmag]{IEEEtran}

\usepackage{subfigure}
\usepackage{tikz}
\usepackage{amsfonts}
\usepackage{amsmath}
\newtheorem{theorem}{Theorem}

\newtheorem{assumption}{Assumption}
\usepackage{amssymb}
\usepackage{hyperref}
\usepackage{algorithm}
\usepackage{algorithmic}
\usepackage{booktabs}
\usepackage{multirow}
\usepackage{makecell}
\usetikzlibrary{automata}
\usetikzlibrary{shapes,arrows}
\newcommand{\tabincell}[2]{\begin{tabular}{@{}#1@{}}#2\end{tabular}} 
\ifCLASSINFOpdf
\else
\fi

\begin{document}
\title{Online Attentive Kernel-Based Temporal Difference Learning} 
\author{Guang Yang, Xingguo Chen,
Shangdong Yang, Huihui Wang, Shaokang Dong, Yang Gao
\IEEEcompsocitemizethanks{
\IEEEcompsocthanksitem G. Yang and X. Chen are with the the Jiangsu Key Laboratory of Big Data Security \&
Intelligent Processing, Nanjing University of Posts and Telecommunications, and
National Engineering Laboratory for Agri-product Quality Traceability, Beijing Technology and Business University, P.R., China \protect (e-mail: guangyang9543@gmail.com; chenxg@njupt.edu.cn).
\emph{(Corresponding author: X. Chen.)}%
\IEEEcompsocthanksitem S. Yang, S. Dong and Y. Gao are with the State Key
Laboratory for Novel Software Technology, Nanjing University, P.R., China \protect (e-mail:
shaokangdong@gmail.com, yangshangdong007@gmail.com, gaoy@nju.edu.cn)
\IEEEcompsocthanksitem H. Wang is with the PCA Lab, Key Lab of Intelligent
Perception and Systems for High-Dimensional Information of Ministry of
Education, and Jiangsu Key Lab of Image and Video Understanding for Social
Security, School of Computer Science and Engineering, Nanjing University of
Science and Technology, P. R., China \protect (e-mail: Huihuiwang@njust.edu.cn).
}
}


\maketitle
\begin{abstract}
With rising uncertainty in the real world, online Reinforcement Learning (RL)
has been receiving increasing attention due to its fast learning capability and
improving data efficiency. However, online RL often suffers from
complex Value Function Approximation (VFA) and catastrophic interference,
creating difficulty for the deep neural network to be applied to an online RL
algorithm in a fully online setting. Therefore, a simpler and more adaptive
approach is introduced to evaluate value function with the kernel-based model.
Sparse representations are superior at handling interference, indicating that
competitive sparse representations should be learnable, non-prior,
non-truncated and explicit when compared with current sparse representation
methods. Moreover, in learning sparse representations, attention mechanisms are
utilized to represent the degree of sparsification, and a smooth attentive
function is introduced into the kernel-based VFA. In this paper, we propose an
Online Attentive Kernel-Based Temporal Difference (OAKTD) algorithm
using two-timescale optimization and provide convergence analysis of our proposed
algorithm. Experimental evaluations showed that OAKTD outperformed several
Online Kernel-based Temporal Difference (OKTD) learning algorithms in addition
to the Temporal Difference (TD) learning algorithm with Tile Coding on public
Mountain Car, Acrobot, CartPole and Puddle World tasks.
\end{abstract}



\IEEEpeerreviewmaketitle

\input{introduction.tex}

\input{background.tex}
\input{method.tex}

\input{convergence.tex}

\input{experiments.tex}

\input{conclusion.tex}

\bibliographystyle{IEEEtran} 
\bibliography{references} 
\end{document}

%% file: introduction.tex
\section{Introduction}

\IEEEPARstart{I}{n} Reinforcement Learning (RL), an agent usually seeks the
optimal policy and learns it to solve a sequential decision-making problem
\cite{sutton1998reinforcement}, and the entire learning process relies heavily
on the interaction between agent and working environment. In practice,
real-world problems are complex and indefinite, often resulting in ineffective
and inefficient computation and learning during long episodic applications
while using offline algorithms. To solve these problems, online RL has been
proposed, a solution gaining recent widespread attention \cite{liu2019sparse},
especially online RL methods for fully online updating which rely on one
transition\cite{liu2019utility}, 
which is current and limited transition or experience, e.g., Temporal Difference
(TD) learning \cite{sutton1988learning}.
Online RL can result in faster learning through bootstrapping and improved
computational and data efficiency because it focuses on learning from real-time
samples which are encountered frequently. However, online RL implementation
still needs to consider the following issues:
\begin{itemize}
  \item Online RL requires a function approximation that can work effectively in
  online learning rather than offline or batch learning.
  \item Online updating often suffers from catastrophic forgetting or interference.
  \item Online algorithms should possess good convergence guarantees.
\end{itemize}

Value Function Approximation (VFA) is a widely used RL technique for large-scale
or continuous state spaces. In general, it is a mapping of features and related
parameters to state values, which can be divided into linear and nonlinear
methods. As a representative of nonlinear VFA, learning from batch updating and
deep neural networks \cite{lecun2015deep} often fails to solve simple RL tasks
in fully online settings \cite{liu2019sparse}. Compared with deep neural
networks, linear VFA methods can be computed quickly and exhibit elegant
theoretical analysis \cite{sutton2009fast, sun2020finite, lee2019target},
however, their performance often depends on features defined by experts.
Fortunately, kernel-based VFA is an ideal choice for online RL, as it can not
only learn representations adaptively \cite{ormoneit2002kernel}, but also learn
in fully online settings \cite{chen2013online}.

When online RL uses VFA, an update on one transition may change all parameters
of the value function. This issue, referred to catastrophic interference, causes
the update on the current transition to catastrophically interfere with or
forget the updates on previous transitions. Generally, an effective method for
resolving catastrophic interference in online RL is sparse representation 
\cite{liu2019sparse}, where only partial parameters need updating to reduce
global interference. In addition, sparse representation can capture important
state attributes so that the agent can obtain an accurate value evaluation.

Traditional sparse representation methods such as Tile Coding
\cite{sutton1996generalization} and n-tuple networks \cite{krawiec2011learning},
have been successfully applied in linear VFA algorithms, but they are
artificially predefined and limited by the curse of dimensionality. For neural 
networks, there are certain methods that can be used for sparse
representation: ($i$) Rectified Linear Units (ReLU)
\cite{nair2010rectified} is learnable and imply sparsity, but provides no such
guarantees. To further promote sparsity, some heuristics have been
proposed to reduce catastrophic interference. ($ii$) Dropout randomly masks part
of the activation function during training \cite{srivastava2014dropout}.
($iii$) $k$-sparse autoencoders only retain the
top-$k$ nodes largest activations \cite{makhzani2013k}. ($iv$) Winner-Take-All
autoencoders keep the first $k\%$ activations of each layer in the training
\cite{makhzani2015winner}. Although they suited for high-dimensional
inputs, they are problematic, as they have a tendency truncate potentially
significant outputs, possibly producing insufficiently sparse
representations \cite{liu2019utility}. Regularization
strategy is an alternative approach: ($i$) $L_1-$regularization refers to
feature selection, which sets all unimportant features to zero
\cite{park2007l1} and is also a truncated sparse representation. ($ii$)
$L_2-$regularization is relatively smooth \cite{girosi1995regularization},
and implicitly sparse like ReLU.
($iii$) Distributional Regularizers estimate the neural
network output to a sparse distribution, which effectively ensures smooth
sparsity \cite{liu2019utility}, however it must rely on a prior distribution. As
shown in Table \ref{CSR},
\begin{table}
\caption{Comparison of Various Sparse Representations on four aspects: \upshape
``Learnable'' refers to the ability to adaptively learn a sparse representation without
artificial construction; ``Non-prior'' means that no prior knowledge or
hyperparameters are required; ``Non-truncated'' signifies a continuous or smooth
reduction (not discretization); ``Explicit'' means that the value function is
represented by an explicit sparse mechanism.}
\label{CSR}
\begin{tabular}{|p{4.0cm}|p{0.6cm}|p{0.4cm}|p{0.9cm}|p{0.8cm}|}
\hline 
Sparse Representations & \tabincell{c}{Learn- \\ able}& \tabincell{c}{Non- \\
prior} & \tabincell{c}{Non- \\ truncated} & Explicit
\\
\hline  
Tile Coding \cite{sutton1996generalization}  &  &  & & $\surd$\\
\hline 
n-tuple networks \cite{krawiec2011learning} & &  &  & $\surd$\\
\hline 
ReLU \cite{nair2010rectified} & $\surd$ & $\surd$ &  & \\
\hline 
Dropout \cite{srivastava2014dropout} & &  &  & $\surd$ \\
\hline 
$k$-sparse autoencoders \cite{makhzani2013k} & & & & $\surd$ \\
\hline 
Winner-Take-All autoencoders \cite{makhzani2015winner}  & & & & $\surd$ \\
\hline 
$L_1$ regularization \cite{park2007l1} & $\surd$ &  &  & $\surd$ \\
\hline 
$L_2$ regularization \cite{girosi1995regularization}  & $\surd$ &  & $\surd$ & \\
\hline 
Distributional Regularizers \cite{liu2019utility}  & $\surd$ &  & $\surd$ & $\surd$\\
\hline 
selective function \cite{chen2013online} & &  &  & $\surd$ \\
\hline 
attentive function \cite{xu2015show} & $\surd$ & $\surd$ &
$\surd$ & $\surd$\\
\hline 
\end{tabular}
\end{table}
we can conclude that a good sparse representation needs to satisfy four
characteristics: learnable, non-prior, non-truncated and explicit. To learn an
efficient sparse representation, we further propose attentive kernel-based VFA.

Attention mechanism can smoothly extract important features with normalized
weights, which has been successfully applied in visual and machine translation
tasks \cite{mnih2014recurrent, bahdanau2015neural}. In terms of emphasising
local information, attention mechanism is consistent with sparse representation.
Generally, attention mechanism encoding methods can be separated into
hard-attention and soft-attention \cite{xu2015show}, which corresponds to two
different sparsifications: truncated and non-truncated. Previously, Online
Selective Kernel-based Temporal Difference (OSKTD) learning artificially defines
a selective function to provide a truncated sparse representation
\cite{chen2013online}, however it lacks adaptability and smoothness. To learn a
non-truncated sparse representation, we focused on soft-attention and used the
attentive function as the degree of sparsification. In addition, to guarantee
stability of the online RL, we introduced two-timescale optimization
\cite{chung2018two} based on TD and propose Online Attentive Kernel-Based
Temporal Difference (OAKTD) learning. Furthermore, we analyzed the convergence
of OAKTD and compared it with
existing methods on several control tasks. The experimental results verify that
OAKTD is the best performing method.

The rest of this paper is organized as follows. Section II gives an
introduction of the Markov Decision Process, kernel-based VFA and online
dictionary construction, and selective kernel-based VFA. In Section III, we
propose an attentive kernel-based VFA and derive the OAKTD algorithm. In
Section IV, we provide a stability analysis of OAKTD. Section V presents experimental settings, 
results and analysis. Finally, conclusions and
future work are stated in Section VI.

\begin{table}
\centering
\caption{Notations.}
\label{notation}
\begin{tabular}{ll}
\hline
\hline
Symbol  & Meaning \\
\hline
$s, s'$   & states\\
$|s|$   & dimensionality of state\\
$a$   & an action  \\
$r$   & a reward  \\
$\gamma$   & discount rate  \\
$\mathcal{S}$  & set of all nonterminal states \\
$|\mathcal{S}|$  & number of elements in set $\mathcal{S}$ \\
$\mathcal{R}$  & reward function \\
$\mathcal{A}$  & set of all actions  \\
$\mathcal{D}$  & diagonal matrix of state distribution \\
$\pi$  & policy (decision-making rule) \\
$t$  & discrete time step \\
$T$  & final time step of an episode\\
$S_t, \hat{S}_t$  & states at time $t$ \\
$R_t$  & reward at time $t$ \\
$\alpha_t, \beta_t$   & step-size at time $t$  \\
$V^{\pi}(s)$  & value of state $s$ under policy $\pi$ (expected return) \\
$V^{*}(s)$  & value of state s under the optimal policy \\
$\theta, \theta_t, \bar{\theta},\bar{\theta}_t$  & weights underlying an
approximate value function \\
$w, w_t$  & parameters of attentive function \\
$\bar{w}, \bar{w}_t$  & parameters of attentive kernel-based VFA, \\
                      & $\bar{w}^{\top}=(w^\top,\bar{\theta}^{\top})$, 
                      $\bar{w}_t^{\top}=(w_t^\top,\bar{\theta}_t^{\top})$\\						
$V_{\theta}(s),V_{\theta, w}(s)$  & approximate value of state $s$ given
parameters $\theta, w$ \\
$U_t$ & target for estimate at time $t$ \\
$\phi(s)$  & feature vector of state $s$ \\
$\phi_w(s)$  & feature vector of state $s$ given parameter $w$ \\
$D$  & a fixed dictionary \\
$|D|$  & number of dictionary vectors \\
$D^*$  & stable dictionary \\
$|D^*|$  & number of stable dictionary vectors\\
$s_i$  & an element in dictionary $D$ \\
$k(s,s_i)$  & kernel function \\
$\beta(s,s_i)$  & selective function \\
$\mu_1, \mu_2$  & thresholds \\
$a_w(s,s_i)$  & attentive funtion given parameter $w$\\
$\langle\cdot\rangle$  & tuple \\
$\subset$  & subset of \\
$\in$  & is an element of \\
$:=$  & equality relationship that is true by definition\\
$W$  &  a compact and convex subset $W\subset \mathbb{R}^{|s|+|D|}$ \\
$\Gamma^W$  & a mapping that projects its argument into subset $W$ \\
$\delta_t$  & temporal difference error at $t$ \\
$||\cdot||$  & $2$-norm \\
$||X||_{\mathcal{D}}$  & $\sqrt{X^\top \mathcal{D} X}$ \\
$\nabla_{\theta_t}V_{\theta}(S_t)$  & Jacobian of $V_{\theta}(S_t)$ at $\theta =
\theta_t$ \\
\specialrule{0em}{1pt}{1pt}
\hline
\hline
\end{tabular}
\end{table}

%% file: background.tex
\section{Background}
\subsection{Markov Decision Process}
Consider a discounted Markov Decision Process (MDP) $\langle
\mathcal{S},\mathcal{A},P,\mathcal{R},\gamma \rangle$, where $\mathcal{S}$ is a
state space, $\mathcal{A}$ is a finite action space, $\mathcal{R}:S\times A
\times S \to \mathbb{R}$ is a reward function, $P:\mathcal{S}\times
\mathcal{A}\times \mathcal{S}\to [0,1]$ is a state transition function and $\gamma\in(0,1)$ is a discount factor \cite{sutton1998reinforcement}.
A policy is a mapping $\pi:\mathcal{S}\times\mathcal{A}\to[0,1]$, which
defines for each action the selection probability conditioned on the state.
The return at time $t$ is defined as the discounted sum of rewards, which is then defined as
\begin{equation*}
G_t := \sum_{i=1}^{\infty}\gamma^{i-1}R_{t+i}.
\end{equation*}
The infinite-horizon discounted value function given policy $\pi$ is  
$V^{\pi}:\mathcal{S}\to \mathbb{R}$, which is then defined as
\begin{equation*}
V^{\pi}(s) := \mathbb{E}[G_t|S_t=s,\pi].
\end{equation*}
The goal is to find an optimal policy $\pi^*=\arg\max_{\pi}V^{\pi}(s)$ to
maximize the expectation of the discounted accumulative rewards in a long
period. Under the optimal policy, the optimal value function
$V^*(s)$ satisfies the Bellman optimality equation 
\begin{equation*}
V^*(s) = \max_{a\in \mathcal{A}}\mathbb{E}[r(s,a)+\gamma
V^*(s')|s'\sim P(s,a,\cdot)],
\end{equation*}
where $r(s,a) = \mathbb{E}[R(s,a,s')|s'\sim P(s,a,\cdot)]$.
For the large-scale or continuous state problems. Value Function Approximation
(VFA) is a generalized method. An online RL algorithm requires low
computational complexity. Thus, a common VFA for online RL is linear,
which can be then defined as
\begin{equation}
V_{\theta}(s)=\theta^{\top} \phi(s)=\sum_{i=1}^m \theta_i\phi_i(s),
\end{equation}
where $\phi(s)=(\phi_1(s),\ldots,\phi_m(s))^{\top}\in\mathbb{R}^m$ is a feature
vector, and $\theta=(\theta_1,\theta_2,\ldots,\theta_m)^{\top}\in\mathbb{R}^m$ is a
weight vector. 
Linear VFA can be quickly computed and possess elegant and sophisticated
theoretical analysis, e.g., Temporal Difference with Gradient Correction (TDC)
\cite{sutton2009fast}, Decentralized Temporal Difference
Learning \cite{sun2020finite}, Target-based Temporal Difference Learning
\cite{lee2019target} etc. However, their performances rely heavily on
pre-defined feature functions. In nonlinear VFA, deep neural networks are
effective at extracting state representations, but fail to be applied in online
RL \cite{liu2019sparse}. Another is kernel-based VFA, which will be introduced
in the following subsection.

\subsection{Kernel-based VFA and Online Dictionary Construction}
Kernel-based value function is a memory-based and non-parametric model.
According to Representer Theorem \cite{scholkopf2001generalized}, the value
function can be projected into the infinite-dimensional Reproducing Kernel
Hilbert Spaces (RKHS), which is then defined as
\begin{equation}
V_\theta(s) = \theta^{\top}k(s) = \sum_{i=1}^{\infty} \theta_i k(s,s_i),
\label{infinitekbfa}
\end{equation}
where kernel function $k:\mathcal{S}\times \mathcal{S}\to\mathbb{R}$ is a
continuous, symmetric, and positive-definite function according to Mercer's
theorem, e.g., a Gaussian kernel $k(s,s_i)=\exp(-\frac{||s-s_i||^2}{2\sigma^2})$.
It is easy to find that the value function is impossible to use
an infinite number of pairwise $\langle\theta_i, k(s,s_i)\rangle$ in
(\ref{infinitekbfa}).
Therefore, kernel reduction for $\{\langle\theta_i, k(s,s_i)\rangle\}_{\infty}$
is indispensable in practice. The method of reduction is called dictionary
construction. Kernel-based value function is then defined as
\begin{equation}
V_\theta(s)=\theta^{\top}k(s)=\sum_{s_i\in D}\theta_i k(s,s_i),
\label{kbvfa}
\end{equation}
where $D$ is a dictionary of states, the number of dictionary
vectors $|D|\ll |\mathcal{S}|$, and learning parameters $\theta\in
\mathbb{R}^{|D|}$.
\par

Dictionary construction \footnote{Dictionary construction is also known as
kernel sparsification. To avoid confusion and to distinguish from sparse
representations, we mainly use the term ``dictionary construction'' in this
paper.} is targeted at reducing infinite kernel function space
\cite{liu2009information, chen2013online}. In fully online settings,
Modified Novelty Criterion (MNC) is a competitive method, as it can adaptively
learn representation and provide lower computational complexity $O(n)$ compared
with Approximation Linear Dependence (ALD) \cite{chen2013online,
liu2009information}. The MNC is a distance-based dictionary construction method
and the distance-based condition for a new feature vector $\phi(S_t)$ is
\begin{equation}
d_t=\min_{s_i\in
D_{t-1}}||\phi(s_i)-\phi(S_t)||^2 \leq \mu_1,
\label{MNC_condition}
\end{equation}
where $\mu_1$ is a threshold
parameter. Using the kernel trick, by substituting $k(s_i, S_t)$ to
$\phi(s_i)^{\top}\phi(S_t)$, we can obtain $d_t=\min_{s_i\in D_{t-1}}
(k(s_i,s_i) - 2k(s_i,S_t)+k(S_t,S_t))$.
Specially, we define $d_t=\min_{s_i\in D_{t-1}}(2-2k(s_i,S_t))$ with Gaussian
kernel. The update rule for the sample dictionary $D$ is then defined as
\begin{equation}
D_t = \left\{
\begin{array}{ll}
D_{t-1}, &  \text{if}\quad d_t\leq\mu_1,\\
D_{t-1}\cup \{\phi(S_t)\},&  \text{otherwise}.\\
\end{array}
\right.
\label{updatedictionary}
\end{equation}
Online dictionary construction can ensure the feasibility of optimization with
finite parameters, and enhances the generalization ability. 
However, some issues still exist in kernel-based VFA with online dictionary
construction.

\subsection{Sparse Representations for Kernel-based VFA}
Catastrophic interference is one critical issue of VFA applied in online RL 
\cite{liu2019utility}. To formalize the interference in RL, the pairwise
interference between two different samples has been proposed
\cite{liu2019sparse}, which can be then defined as 
\begin{equation*}
PI(\theta_t, S_t, \hat{S}_t) := \mathcal{L}(\theta_{t+1}, \hat{S}_t|S_t)-
\mathcal{L}(\theta_{t}, \hat{S}_t),
\end{equation*}
where $S_t$ and $\hat{S}_t$ are two different states. $\mathcal{L}(\theta_{t},
\hat{S}_t)$ is the objective function for the state $\hat{S}_t$.
$\theta_{t+1}$ is updated by the gradient based state $S_t$. If $PI$ is positive, the
interference will occur.
Under traditional TD algorithm, a natural loss function is the Mean Squared
Value Error (MSVE), which is then defined as
\begin{equation*}
\mathcal{L}(\theta) =
\text{MSVE}(\theta):=||V^{\pi}-V_{\theta}||^2_{\mathcal{D}},
\end{equation*}
where $\mathcal{D}$ is the diagonal matrix of state distribution. The update
of $\theta$ is 
\begin{equation*}
\theta_{t+1} = \theta_t + \alpha(U_t-V_{\theta_t}(S_t))\nabla_{\theta_t}V_{\theta}(S_t), 
\end{equation*}
where $\alpha$ is a step-size, $U_t$ is target value and
$\nabla_{\theta_t}V_{\theta}(S_t)$ is the Jacobian of $V_{\theta}(S_t)$ at
$\theta=\theta_t$. 
Then, the pairwise interference in RL can be approximated by a Taylor expansion:
\begin{equation*}
\begin{split}
& PI(\theta_t, S_t, \hat{S}_t) \approx(\theta_{t+1}-\theta_{t})\frac{\partial
\mathcal{L}(\theta_{t}, \hat{S}_t)}{\partial \theta_t} \\
= & 2\alpha[U_t-V_{\theta_t}(S_t)][V^{\pi}(\hat{S}_t)-V_{\theta_t}(\hat{S}_t)]
\nabla_{\theta_t}V_{\theta}(S_t)^{\top}\nabla_{\theta_t}V_{\theta}(\hat{S}_t).
\end{split}
\end{equation*}
This equation provides some insights for interference analysis. It is
difficult to determine whether interference occurs, because $V^{\pi}(\hat{S}_t)$ is
unknown. Fortunately, we can avoid interference satisfying the condition
$\nabla_{\theta_t}V_{\theta}(S_t)^{\top} \nabla_{\theta_t}V_{\theta}(\hat{S}_t)
\to 0$. Furthermore, for a linear VFA, we have 
\begin{equation*} 
\nabla_{\theta_t}V_{\theta}(S_t)^{\top}
\nabla_{\theta_t}V_{\theta}(\hat{S}_t) = \phi(S_t)^{\top}\phi(\hat{S}_t), 
\end{equation*}
where $\phi$ is feature vector. The condition can be also written as
$\phi(S_t)^{\top}\phi(\hat{S}_t) \to 0$. Thus, to alleviate
interference, sparse representations can be an effective method to approximate
the condition.

Similarly, in kernel-based VFA, we have 
\begin{equation*}
\nabla_{\theta_t}V_{\theta}(S_t)^{\top}
\nabla_{\theta_t}V_{\theta}(\hat{S}_t)=\phi(S_t)^{\top}\phi(\hat{S}_t)=k(S_t,
\hat{S}_t),
\end{equation*}
by the kernel trick. However, it is difficult for kernel-based VFA to guarantee
an effective and explicit sparse representation in online RL, because the
condition $k(S_t, \hat{S}_t) \to 0$ is hard to be satisfied. 
In order to encourage sparsification of kernel-based VFA, a selective
kernel-based value function is proposed \cite{chen2013online},
\begin{equation*}
V(s)=\theta^{\top}K(s)\beta(s)=\sum_{s_i\in D}\theta_{i}k(s,s_i) \beta(s,s_i),
\label{skbvfa}
\end{equation*}
where $K(s)$ is the diagonal matrix of kernel function $k(s,s_i)$, and
$\beta(s)$ is a distance-based selective function vector with
elements$\beta(s,s_i)$, which is written as
\begin{equation*}
\beta(s,s_i)=\left\{
\begin{aligned}
1, & \quad k(s,s)-2k(s,s_i)+k(s_i,s_i)<\mu_2,\\
0, & \quad \text{otherwise},
\end{aligned}
\right.
\label{betafunction}
\end{equation*}
where $\mu_2$ is a threshold parameter. 
The selective function truncates partial kernels by a distance-based indicator
function and can also be replaced by other sparse representations in structures.
Two simple examples of this will be provided. In dropout, we have
\begin{equation*}
\beta_{\text{dropout}}(s,s_i)=\left\{
\begin{aligned}
1, & \quad \text{with $\eta$ possibility},\\
0, & \quad \text{otherwise},
\end{aligned}
\right.
\label{betafunction}
\end{equation*}
where $0<\eta<1$. Similarly, for ReLU, we have 
\begin{equation*}
\beta_{\text{ReLU}}(s,s_i)=\left\{
\begin{aligned}
1, & \quad k(s,s_i) > b,\\
0, & \quad \text{otherwise},
\end{aligned}
\right.
\label{betafunction}
\end{equation*}
where $b$ is a learnable bias. In online RL, selective kernel-based VFA provides
a generalized framework for combining several existing sparse methods. However,
as mentioned previously, these methods may exhibit problems in their learnable,
non-prior, non-truncated and explicit characteristics. Therefore, we will
introduce the attentive function in greater detail in the next section.

%% file: method.tex
\section{Online Attentive Kernel-Based Temporal Difference Learning}
The highlights of our methods are: ($i$) we designed an attentive kernel-based
value function, ($ii$) we adopted online two-timescale optimization. In the
following part, details of Online Attentive Kernel-Based Temporal Difference
learning are presented.
\subsection{Attentive Kernel-based VFA}
In order to obtain a good sparse representation, we introduce an attentive
function. 
\begin{figure}[ht]
\centering
	\resizebox{7.5cm}{6cm}{
	\begin{tikzpicture}
            \centering
            \tikzstyle{every state}=[draw=black,very thick,minimum size=.8cm,inner sep=.4]
            \node[state,font=\Large] (s) {$s$};
            \node[state,font=\Large] (s_0) at (2.5,3.8) {$s_1$};
            \node[state,font=\Large] (s_1) at (2.5,.6) {$s_2$};
            \node[state,font=\Large] (s_D) at (2.5,-3) {$s_{|D|}$};
            \node[shape=circle,draw=black,very thick,minimum size=1.5cm,inner
            sep=.75, font=\Large] (t1) at (4.5,4.6) {$k(s,s_1)$};
            \node[shape=circle,draw=black,very thick,minimum size=1.5cm,inner
            sep=.75, font=\Large] (t2) at (5.5,3) {$a_w(s,s_1)$};
            \node[shape=circle,draw=black,very thick,minimum size=1.5cm,inner
            sep=.75, font=\Large] (t3) at (4.5,1.4) {$k(s,s_2)$};
            \node[shape=circle,draw=black,very thick,minimum size=1.5cm,inner
            sep=.75, font=\Large] (t4) at (5.5,-.2) {$a_w(s,s_2)$};
            \node[shape=circle,draw=black,very thick,minimum size=1.5cm,inner
            sep=.75, font=\large] (t5) at (4.5,-2.2) {$k(s,s_{|D|})$};
            \node[shape=circle,draw=black,very thick,minimum size=1.5cm,inner
            sep=.75, font=\large] (t6) at (5.5,-3.8) {$a_w(s,s_{|D|})$};
            \node[shape=rectangle,draw=black,very thick,minimum
            size=0.8cm,font=\LARGE] (p1) at (7.5,3.8) {$\times$};
            \node[shape=rectangle,draw=black,very thick,minimum
            size=0.8cm,font=\LARGE] (p2) at (7.5,.6) {$\times$};
            \node[shape=rectangle,draw=black,very thick,minimum
            size=0.8cm,font=\LARGE] (p3) at (7.5,-3) {$\times$};
            \node[shape=rectangle,draw=black,very thick,font=\Large] (sum) at
            (10,0) {$\sum$};
            \node[state,font=\Large] (V) at (11.5,0) {$V_{\theta,w}(s)$};

            \node[font=\large] at (0,.8) {State} ;
            \node[font=\large] at (2.5,4.6) {Dictionary};
            \node[font=\Large] at (9,4.5) {$\theta$};
            \node at (2.5,-1) {$\vdots$};
            \node at (5,-1.05) {$\vdots$};
            \node at (7.5,-1) {$\vdots$};
            \node[font=\Large] at (9,2) {$\theta_{1}$};
            \node[font=\Large] at (9,.5) {$\theta_{2}$};
            \node[font=\Large] at (9,-1.8) {$\theta_{|D|}$};

            \draw[dashed] (1.9,4.3) -- (3.1,4.3)
                          (1.9,4.3) -- (1.9,-3.6)
                          (1.9,-3.6) -- (3.1,-3.6)
                          (3.1,4.3) -- (3.1,-3.6);

            \draw[dashed] (8.4,4.2) -- (9.5,4.2)
                          (8.4,4.2) -- (8.4,-3.8)
                          (8.4,-3.8) -- (9.5,-3.8)
                          (9.5,4.2) -- (9.5,-3.8);

            \path[->, very thick] (s)   edge node [above left] {} (s_0)
                            edge node [above left] {} (s_1)
                            edge node [above left] {} (s_D)
                    (s_0)   edge node [above left] {} (t1)
                            edge node [above left] {} (t2)
                    (s_1)   edge node [above left] {} (t3)
                            edge node [above left] {} (t4)
                    (s_D)   edge node [above left] {} (t5)
                            edge node [above left] {} (t6)
                    (t1)    edge node [above left] {} (p1)
                    (t2)    edge node [above left] {} (p1)
                    (t3)    edge node [above left] {} (p2)
                    (t4)    edge node [above left] {} (p2)
                    (t5)    edge node [above left] {} (p3)
                    (t6)    edge node [above left] {} (p3)
                    (p1)    edge node [above left] {} (sum)
                    (p2)    edge node [above left] {} (sum)
                    (p3)    edge node [above left] {} (sum)
                    (sum)   edge node [above left] {} (V);

        \end{tikzpicture}
	}
\caption{Attentive Kernel-based VFA. $D=\{s_1, s_2,
\cdots, s_{|D|}\}$ is a fixed dictionary, and the multiplication of $k(s,\cdot)$
and $a_w(s,\cdot)$ is used as a sparse representation.}
\label{model}
\end{figure}
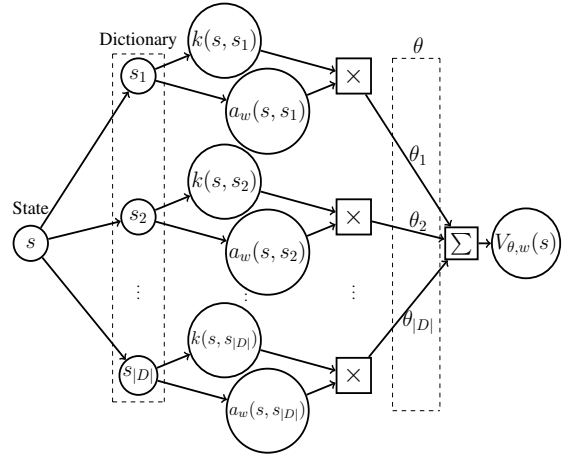
As shown in Fig. \ref{model}, based on the framework of sparse representation
for kernel-based VFA, we have attentive kernel-based value
function\footnote{For convenience of expression, we use $V_{\theta,w}$
to replace $V_{\theta,w, D}$, which is online attentive kernel-based VFA about
dictionary $D$. In subsequent theoretical proofs, we take dictionary
convergence into account and re-add the dictionary symbol, e.g.,
$\Phi_{w,D}$.}
\begin{equation}
V_{\theta,w}(s)=\theta^{\top}K(s)a_w(s)=\sum_{s_i\in
D}\theta_{i}k(s,s_i)a_w(s,s_i),
\end{equation}
where $K(s)$ is the diagonal matrix of kernel function $k(s,s_i)$, parameters
$w$ and $\theta$ are learning parameter vectors, and $a_w(s)$ is the attentive
function vector with elements $a_w(s,s_i)$. $D$ is a dictionary constructed
online by MNC.

Our model uses the attentive function as the degree of sparsification. Based on
the generalized attention model, we defined a novel score function based on
$L_1$ Norm
\begin{equation}
\begin{split}
e(s,s_i) &= w^{\top} \psi(s,s_i) , \\
\end{split}
\end{equation}
where the learning parameter $w\in\mathbb{R}^{|s|}$, and
$\psi(s,s_i)\in\mathbb{R}^{|s|}$ is the vector with the
$k$-th element $|\chi_k(s)-\chi_k(s_i)|$, where $\chi_k(s)$ is the $k$-th
element of state $s$.
After normalization by the softmax function, the attentive function is
then defined as
\begin{equation}
a_{w}(s,s_i) = \frac{\exp(w^{\top}\psi(s,s_i))}
{\sum_{s_j\in D}\exp(w^{\top}\psi(s,s_j))}.
\end{equation}
Attentive kernel-based VFA is nonlinear, which may lead to algorithm
instability. Thus, we introduced two-timescale optimization to alleviate this
issue in the next subsection.

\subsection{Two-Timescale Optimization}
\begin{figure}[ht]
\centering 
\includegraphics[scale=0.3]{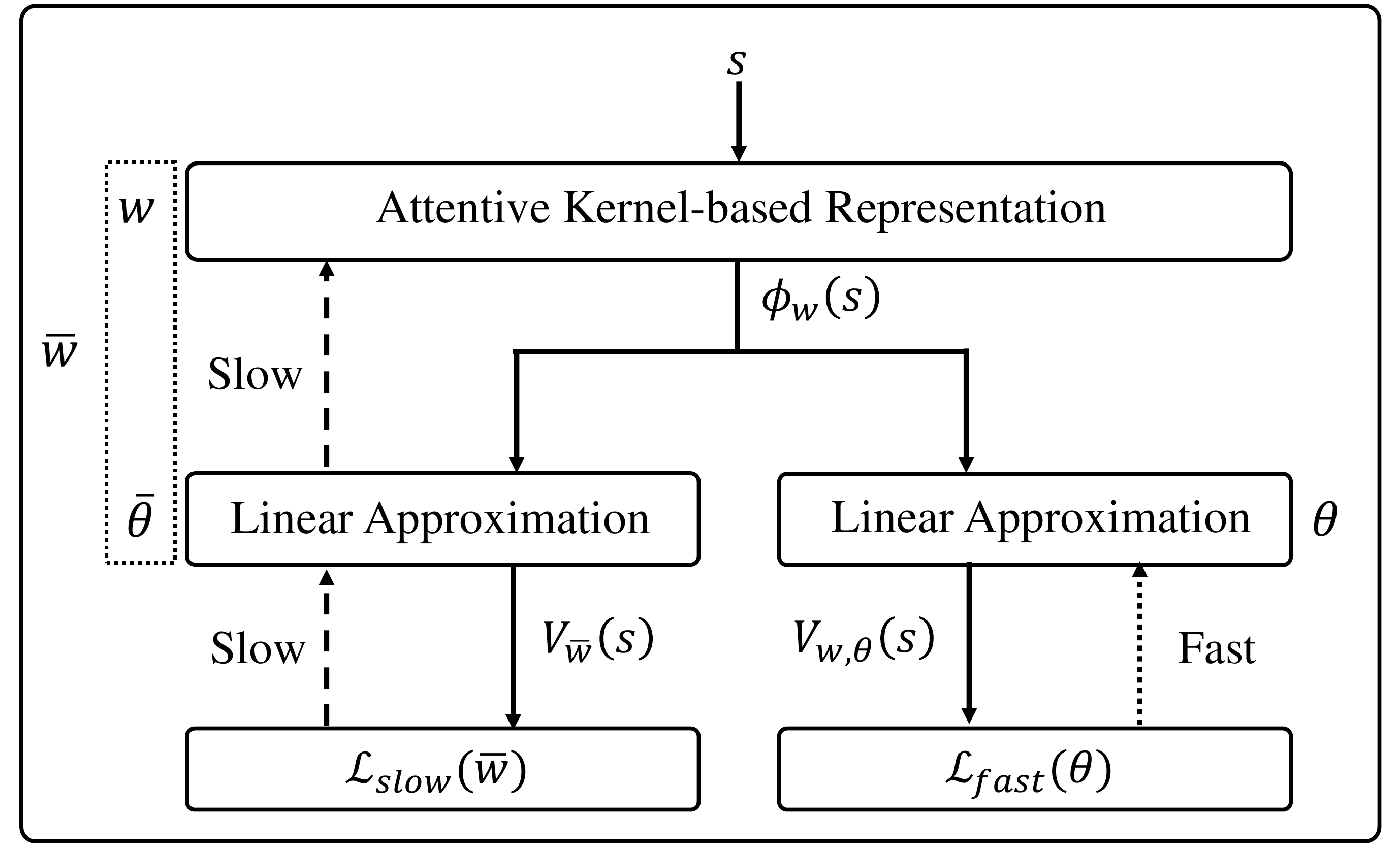}
\caption{Two-timescale optimization. The left is a slow learning
process, where $\bar{\theta}$ is an auxiliary learning parameter and
$\bar{w}^{\top}=(w^\top,\bar{\theta}^{\top})$. The right is a fast learning
process and the dotted line shows the loss backpropagation.}
\label{TTO}
\end{figure}

To guarantee stability of attentive kernel-based VFA, we adopt two-timescale
optimization \cite{chung2018two}, which in process was divided into slow
and fast parts.
As shown in Fig. \ref{TTO},
the slow part uses an auxiliary parameter $\bar{\theta}$ to learn attentive
kernel-based sparse features $\phi_w$ and the fast part approximates a linear
value function $V_{w,\theta}$ with the learned features. In structure,
two-timescale optimization splits attentive kernel-based VFA into two parts:
attentive sparse representation and linear approximation. Then, attentive
kernel-based value function can be also written as
\begin{equation*}
    V_{\theta,w}(s)=\theta^{\top}\phi_{w}(s),
\end{equation*}
where $\phi_{w}(s)=K(s)a_w(s)$ is attentive kernel-based features. In slow
process, to obtain a stable attentive kernel-based representation learning, we
use Mean Squared Bellman Error (MSBE) as objective function, which can be written as
\begin{equation*}
\begin{split}
\text{MSBE}(\bar{w}) &= ||\mathbb{T}V_{\bar{w}}-V_{\bar{w}}||_\mathcal{D}^2 \\
&=\sum_{s\in S}d(s)(\mathbb{E}[\delta_t|S_t=s])^2,
\end{split}
\end{equation*}
where $\bar{w}^{\top}=(w^\top,\bar{\theta}^{\top})$ are learning parameters,
$\mathbb{T}$ is known as Bellman operator, $\delta_t=R_{t+1} + \gamma
V_{\bar{w}_t}(S_{t+1})-V_{\bar{w}_t}(S_{t})$ is one-step TD error, $d(s)$ is
distribution of state $s$, $\sum_{s\in S}d(s)=1$, and $\mathcal{D}$ is a
diagonal matrix of $d(s)$. Generally, in residual-gradient algorithm, Bellman
error for a state is expected TD error in that state \cite{baird1995residual}.
To get an unbiased estimate for MSBE, double sampling of subsequent states is
required. However, in online interaction with an environment, this would seem be
impossible. Thus, we adopt naive residual-gradient based on the Mean Squared
Temporal Difference Error (MSTDE), $\sum_{s\in
S}d(s)\mathbb{E}[\delta_t^2|S_t=s]$ \cite{sutton1998reinforcement}, and the
objective function in slow process is defined as
\begin{equation*}
\begin{split}
\mathcal{L}_{slow}(\bar{w})
&=\sum_{s\in S}d(s)\big(r+\gamma V_{\bar{w}}(s')-V_{\bar{w}}(s)\big)^2,
\end{split}
\end{equation*}
For a nonlinear VFA, the update of slow part is generally by a projection
operator described as
\begin{equation}
    \bar{w}_{t+1}=
    \Gamma^{W}\Big(\bar{w}_{t}+\alpha_t\delta_t(\nabla_{\bar{w}_{t}}V_{\bar{w}}(S_t)-
    \gamma\nabla_{\bar{w}_{t}}V_{\bar{w}}(S_{t+1}))\Big),
    \label{slow}
\end{equation}
where $\alpha_t$ is a step-size, 
$\nabla_{\bar{w}_{t}}V_{\bar{w}}(s)\in\mathbb{R}^{|s|+|D|}$ is the
Jacobian of $V_{\bar{w}}(s)$ at $\bar{w}=\bar{w}_t$, and $\Gamma^{W}$ is a
mapping that projects its argument into an appropriately chosen compact convex
subset $W\subset\mathbb{R}^{|s|+|D|}$ with a smooth boundary  \footnote{The
purpose of this projection is to prevent the parameters to diverge in the
initial phase of the nonlinear approximation. It is very likely that no
projections will take place at all if one selects $W$ large enough
\cite{Hamid2009convergent}. One of the main reason for the projection is to
facilitate convergence analysis.}.

Furthermore, we deduce the update formula based on gradient in terms of their
respective components. For a linear FA, we have
\begin{equation}
\begin{split}
\bar{\theta}_{t+1}
&=\Gamma^{W}\Big(\bar{\theta}_{t}+\alpha_{t}\delta_{t}(\nabla_{\bar{\theta}_t}V_{\bar{w}}(S_t)-
\gamma  \nabla_{\bar{\theta}_t}V_{\bar{w}}(S_{t+1})) \Big)\\
&=\Gamma^{W}\Big(\bar{\theta}_{t}+\alpha_{t}\delta_{t}(K(S_t)a_{w_t}(S_t)-\gamma
K(S_t)a_{w_t}(S_{t+1})) \Big).
\label{updatetheta}
\end{split}
\end{equation}
In a representation learning, we can obtain the same update of $w_{t+1}$, which
is written as 
\begin{equation}
\begin{split}
w_{t+1} &=
\Gamma^{W}\Big(w_{t}+\alpha_{t}\delta_{t}(\nabla_{w_t}V_{\bar{w}}(S_t)- \gamma
\nabla_{w_t}V_{\bar{w}}(S_{t+1})) \Big) \\
&=\Gamma^{W}\Big(w_{t}+\alpha_{t}\delta_{t}(\nabla_{w_t}\phi^{\top}_{w}(S_t)-\gamma
\nabla_{w_t}\phi^{\top}_{w}(S_{t+1}))\bar{\theta}_t \Big),\\
\end{split}
\label{w0}
\end{equation}
where $\nabla_{w_t}\phi_{w}^{\top}(s)\in\mathbb{R}^{|s|\times|D|}$ is the
Jacobian of $\phi^{\top}_{w}(s)$ at $w=w_t$, and
$\nabla_{w_t}\phi_{w}(s,s_i)=k(s,s_i)\nabla_{w_t}a_{w}(s,s_i)$ is the column
corresponding to $s_i$. For convenience, we define
\begin{equation*}
\begin{split}%
&\dot{a}_{w_t}(s,s_i):= \nabla_{w_t} a_{w_t}(s,s_i)\\
=&\frac{\nabla_{w_t} e(s,s_i)\exp(e(s,s_i))}
{\sum_{s_j\in D}\exp(e(s,s_j))}- \\
&\frac{\exp(e(s,s_i))\sum_{s_j\in D}\nabla_{w_t}
e(s,s_j)\exp(e(s,s_j))}{\sum_{s_j\in D}^2 \exp(e(s,s_j))} \\
=& a_{w_t}(s,s_i)\Big[\psi(s,s_i)-\frac{\sum_{s_j\in
D}\psi(s,s_j)\exp(e(s,s_j))} {\sum_{s_j\in
D}\exp(e(s,s_j))}\Big]\\
= &a_{w_t}(s,s_i)[\psi(s,s_i) -
\sum_{s_j\in D}a_{w_t}(s,s_j)\psi(s,s_j)].
\end{split}
\end{equation*}
Then, the updating of $w_{t+1}$ in (\ref{w0}) can be rewritten as
\begin{equation}
\begin{split}
    w_{t+1}= \Gamma^{W}\Big(w_{t}+\beta_t\delta_t 
    & \sum_{s_i\in D}\bar{\theta}_{i,t}\big(
    k(S_t,s_i)\dot{a}_{w_t}(S_t,s_i)- \\
    & \gamma k(S_{t+1},s_i)\dot{a}_{w_t}(S_{t+1},s_i)
    \big)\Big).
\end{split}
\label{updatew}
\end{equation}

After attentive kernel-based representation learning, a linear state value
estimate is available. In fast process, a natural objective function can
use the Mean Squared Value Error (MSVE) which measures the difference between
the approximate value $V_{\theta, w}$ and the true value $V^{\pi}$
\cite{sutton1998reinforcement}, which is defined as
\begin{equation*}
    \begin{split}
        \mathcal{L}_{fast}(\theta) &= ||V^{\pi}-V_{\theta,w}||^2_{\mathcal{D}}\\
        &=\sum_{s\in\mathcal{S}}d(s)[V^{\pi}(s)-V_{\theta,w}(s)]^2.
    \end{split}
\end{equation*}
Because $V^{\pi}$ is unknown, a prototypical
semi-gradient method uses $U_t=R_{t+1}+\gamma V_{\theta_{t},w_{t}}(S_{t+1})$ as
its target in place of $V^{\pi}(S_t)$ after calculating its gradient, and the
update of the fast is defined as
\begin{equation}
\begin{split}
\theta_{t+1}
&=\theta_{t}+\beta_{t}(U_t-V_{\theta_t, w_t}(S_t))\nabla_{\theta_t}V_{\theta,w}(S_t)\\
&=\theta_{t}+\beta_{t}\delta_{t}K(S_t)a_{w_t}(S_t),
\label{fast}
\end{split}
\end{equation}
where $\beta_t$ is a fast learning step-size, satisfying $\frac{\alpha_t}{\beta_t}\to 0$ as $t\to\infty$.

\subsection{Algorithm Description}
Our proposed OAKTD is an online learning algorithm, which aims to enhance
accuracy and stability of kernel-based algorithms. The main steps of OAKTD include three
aspects:
(i) Online dictionary construction; 
(ii) Attentive kernel-based representation learning;
(iii) Linear value function evaluation.

Our OAKTD is based on traditional RL paradigm and the classical framework of
on-policy TD. The MNC-based dictionary construction is a distance-based method
with a finite number of dictionary vectors which appeals to Theorem $2.1$,
Chapter $2.2$ of Engle \cite{engel2004kernel, xu2007kernel}. Additionally, we
can adjust threshold $\mu_1$ and enhance the exploration to make the dictionary
converge at a very fast speed. As result, the dictionary convergence time was
short and negligible.
Then, we calculated the computational complexity in OAKTD using $n+n^2+nm$,
where $n=|D^*|$ is  is the number of stable dictionary vectors and $m=|s|$ is
state's dimension.
In reality, the state’s dimension is generally far less than the dictionary size
$m\ll n$, the real computational complexity of each update of OAKTD is $O(n^2)$.
In summary, pseudocode for the OAKTD is summarized as Algorithm \ref{oaktd}.

\begin{algorithm}
\caption{Online Attentive Kernel-Based Temporal Difference Learning}
\begin{algorithmic}[1]
\STATE Input: the policy $\pi$ to be evaluated
\STATE Algorithm parameter: $\alpha$, $\beta$, $\epsilon$, $\mu_1$
\STATE Initialize parameters $D=\varnothing$, $\theta$, $\bar{\theta}$, $w$
\STATE \textbf{Loop} for each episode:
	\STATE \quad initialize state $s$
	\STATE \quad \textbf{Loop} for each step of episode:
	\STATE \quad \quad Sample $a\sim \pi(s,\cdot)$
	\STATE \quad \quad Take action $a$, observe $r$, $s'$
	\STATE \quad \quad Update dictionary $D$ according to
	(\ref{updatedictionary})
	\STATE \quad \quad Update $\bar{\theta}$, $w$ according to 
	(\ref{updatetheta}), (\ref{updatew}) 
	\STATE \quad \quad Update $\theta$ according to
	(\ref{fast}) 
	\STATE \quad \quad Update learning rate $\alpha$, $\beta$ and greedy rate
	$\epsilon$ \STATE \quad \quad $s\leftarrow s'$
	\STATE \quad \textbf{Until} s is terminal
\end{algorithmic}
\label{oaktd}
\end{algorithm}











%% file: convergence.tex
\section{The Stability Analysis of OAKTD}

In this section, we analyze 
the convergence of online dictionary
construction, and the convergence of two-timescale optimization under a fixed
dictionary assumption and other standard assumptions.

\begin{assumption}
The kernel function $k$ in dictionary construction is a Lipschitz
continuous Mercer kernel and state space $\mathcal{S}$ is a compact subset of a
Banach space.
\label{A1}
\end{assumption} 

\begin{theorem}[Convergence of MNC-based Dictionary Construction]
Let Assumption \ref{A1} hold. Under MNC-based dictionary construction
procedure, for any training sequence $\{s_i\}\in
\mathcal{S}(i=1,2,\dots,\infty)$ and any $\mu_1 > 0$, the number of
dictionary vectors is finite, and the dictionary sequence $\{D_t\}_{t\in\mathbb{N}}$
converges to a stable dictionary $D^*\subset\mathcal{S}$.
\label{T1}
\end{theorem}

\begin{IEEEproof}
According to Assumption \ref{A1}, we claim that for any training sequence
$\{s_i\}\in \mathcal{S}(i=1,2,\dots,\infty)$ and for any $\mu_1 > 0$, the number of
dictionary vectors is finite, and the dictionary sequence
$\{D_t\}_{t\in\mathbb{N}}$ converges to a stable dictionary
$D^*\subset\mathcal{S}$.
According to the MNC-based dictionary construction (\ref{updatedictionary}), it
is easy to find that any two elements $\phi(s_i)$ and $\phi(s_j)$ in the
dictionary are $\sqrt{\mu_1}$-separated, satisfying
$||\phi(s_i)-\phi(s_j)||>\sqrt{\mu_1}$. Then, based on the analysis given by
Theorem $2.1$ and Proposition $2.2$, Chapter $2.2$ of Engle
\cite{engel2004kernel, xu2007kernel}, the claim follows.
\end{IEEEproof}

We consider a function $\Gamma:U\subseteq{\mathbb{R}^{d_1}}\to
X\subseteq{\mathbb{R}^{d_2}}$ is Frechet differentiable at $x\in U$, i.e., 
there exists a bounded linear operator
$\hat{\Gamma}_x:\mathbb{R}^{d_1}\to\mathbb{R}^{d_2}$ such that the limit
\begin{equation*}
    \hat{\Gamma}_x(y)=\lim_{\epsilon\downarrow 0} \frac{\Gamma(x+\epsilon y)-x}{\epsilon}.
\end{equation*}
Recall that $\Gamma^{W}$ is the projection onto a prescribed compact and
convex subset $W\subset\mathbb{R}$, where $\Gamma^{W}(x)=x$,
for $x\in\mathring{W}$, $\mathring{W}$ is the interior of $W$,
while for $x\notin\mathring{W}$, it is the nearest point in $W$
w.r.t. the Euclidean distance $\Gamma^{W}(x)=\arg\min_{x'\in
W}||x'-x||$. Then, the above limit exists when the boundary
$\partial W$ of $W$ is smooth. 
Further, for $x\in\mathring{W}$, we have
\begin{equation*}
    \begin{split}
        \hat{\Gamma}^{W}_x(y) 
        & =\lim_{\epsilon\to 0}\frac{\Gamma^{W}(x+\epsilon y)-x}{\epsilon}
        = \lim_{\epsilon\to 0}\frac{x+\epsilon y-x}{\epsilon} =y.
    \end{split}
\end{equation*}
i.e., $\hat{\Gamma}^{W}_x(\cdot)$ is an identity map for
$x\in\mathring{W}$.

\begin{assumption}
The Markov chain induced by the given policy $\pi$ is ergodic, i.e., aperiodic and
irreducible.
\label{A2}
\end{assumption}

\begin{assumption}
Given a realization of the transition dynamics of the MDP in the form of a
sample trajectory $\mathcal{O}_{\pi}=\{S_0,A_0,R_1,A_1,R_2,S_2,\dots\}$, where
the initial state $S_0\in \mathcal{S}$ is chosen arbitrarily, while the action
$A_t\sim \pi(S_t,\cdot)$, the transitioned state $S_{t+1}\sim P(S_t,A_t,\cdot)$
and the reward $R_{t+1}=r(S_t,A_t,S_{t+1})$.
\label{A3}
\end{assumption}

\begin{assumption}
The pre-determined, deterministic, step-size sequences
$\{\alpha_t\}_{t\in\mathbb{N}}$ and $\{\beta_t\}_{t\in\mathbb{N}}$ satisfy
$\alpha_t,\beta_t\in(0,1]$, $\sum_{t=0}^{\infty}\alpha_t=\sum_{t=0}^{\infty}\beta_t=\infty$,
$\sum_{t=0}^{\infty}\alpha_t^2<\infty$, $\sum_{t=0}^{\infty}\beta_t^2<\infty$
and $\lim_{t\to\infty}\frac{\alpha_t}{\beta_t} = 0$. 
\label{A4}
\end{assumption}

Note that 
$\lim_{t\to\infty}\frac{\alpha_t}{\beta_t}= 0$ implies that
$\{\alpha_t\}$ converges to $0$ relatively faster than $\{\beta_t\}$. The
purpose of using different learning rates is to form a quasi-stationary estimate
\cite{borkar1997stochastic}. When viewed from the faster
timescale recursion, the slower timescale recursion seems quasi-static. When
viewed from the slower timescale, the faster timescale recursion seems
 equilibrated. While analyzing the asymptotic behaviour of the relatively
faster timescale stochastic recursion, it is analytically admissible to
consider the slow timescale stochastic recursion to be quasi-stationary
\cite{sutton2009fast, chung2018two}.

\begin{theorem}[Convergence of Two-Timescale Optimization] 
Let Assumptions \ref{A2}-\ref{A4} hold. Let $D\subset\mathcal{S}$ is a fixed
dictionary. Let $\bar{w}^{\top}=(w^\top,\bar{\theta}^{\top})$ and $\Gamma^{W}$
is a projection to a compact, convex subset $W\subset\mathbb{R}^{|s|+|D|}$
with smooth boundary. Let $\Gamma^{W}$ be Frechet differentiable and
$\hat{\Gamma}_{\bar{w}}^{W}(-\frac{1}{2}\nabla{\mathcal{L}_{slow}})(\bar{w})$
be Lipschitz continuous. Let $\mathcal{W}$ be the set of asymptotically stable
equilibria of the following Ordinary Differential Equation (ODE) contained
inside $W$:
\begin{equation}
\frac{d}{dt}\bar{w}(t)=\hat{\Gamma}^{W}_{\bar{w}(t)}(-\frac{1}{2}\nabla_{\bar{w}}
\mathcal{L}_{slow})(\bar{w}(t)),
\end{equation}
where $\bar{w}(0)\in\mathring{W}$ and $t\in\mathbb{R}_+$. Then the
stochastic sequeue $\{\bar{w}_t\}_{t\in\mathbb{N}}$ generated by natural
residual-gradient algorithm within the two-timescale optimization converges
almost surely to $\mathcal{W}$. Furthermore, the stochastic sequence
$\{\theta_t\}_{t\in\mathbb{N}}$ generated by semi-gradient algorithm within the
two-timescale setting converges almost surely to the limit $\theta^*$, which satisfies
\begin{equation}
	\Phi_{\bar{w}^*,D}\theta^*=\Pi_{\bar{w}^*,D}\mathbb{T}(\Phi_{\bar{w}^*,D}\theta^*),
\end{equation}
where $\bar{w}^*\in\mathcal{W}$, $\Phi_{\bar{w}^*,D}$ is the
attentive kernel-based feature matrix with the $\phi_{\bar{w}^*,D}(s)$ as its
rows, and $\Pi_{\bar{w}^*,D}$ is a projection operator according to
$\Pi_{\bar{w}^*}V=\arg\min_{\bar{V}\in\mathcal{F}}||\bar{V}-V||_{\mathcal{D}}^2$
with $\mathcal{F}=\{\Phi_{\bar{w}^*,D}\theta\}$.
\end{theorem}

\begin{IEEEproof}
The proof is very similar to that given by two-timescale networks
\cite{chung2018two}. In particular, we assume that the dictionary is
fixed, which is to avoid the disturbance caused by dictionary construction
\cite{teschl2012ordinary}.

\textbf{In the slow optimization process}, to analyze the behaviour of the slow
optimization, we apply the ODE-based analysis of stochastic recursive algorithms
\cite{kushner2012stochastic, borkar2009stochastic}. The ODE-based analysis is
elegant, conclusive and further guarantees that limit points of the stochastic
recursion will almost surely belong to the compact internally connected
chain transitive invariant set of the equivalent ODE.
 
Define the filtration $\{\mathcal{F}_t\}_{t\in\mathbb{N}}$, a family of
increasing natural $\sigma-$fields, where
$\mathcal{F}_t\doteq\sigma(\{\bar{w}_i, S_i, R_i;0<i<t\})$. Then, we recall the
previous projected stochastic recursion updating $\bar{w}$, which can be rewritten as
\begin{equation}
\bar{w}_{t+1}=\Gamma^{W}\Big(\bar{w}_t+\alpha_t\big(h(\bar{w}_t)+M_{t+1}+\mathnormal{l}_t\big)\Big),
\end{equation}
where $h(\bar{w}_t)\doteq
\mathbb{E}\big[\delta_t\big(\nabla_{\bar{w}_{t}}V_{\bar{w}}(S_t)-
\gamma\nabla_{\bar{w}_{t}}V_{\bar{w}}(S_{t+1})\big)\big]$, the noise term
$M_{t+1}\doteq \delta_t\big(\nabla_{\bar{w}_{t}}V_{\bar{w}}(S_t)-
\gamma\nabla_{\bar{w}_{t}}V_{\bar{w}}(S_{t+1})\big) -
\mathbb{E}\big[\delta_t\big(\nabla_{\bar{w}_{t}}V_{\bar{w}}(S_t)-
\gamma\nabla_{\bar{w}_{t}}V_{\bar{w}}(S_{t+1})\big)|\mathcal{F}_t\big]$ and the
bias $\mathnormal{l}_t\doteq \mathbb{E}\big[\delta_t\big(\nabla_{\bar{w}_{t}}V_{\bar{w}}(S_t)-
\gamma\nabla_{\bar{w}_{t}}V_{\bar{w}}(S_{t+1})\big)|\mathcal{F}_t\big]-h(\bar{w}_t)$.

Further, 
\begin{equation}
\begin{split}
\bar{w}_{t+1}
&=
\bar{w}_t+\alpha_t\frac{\Gamma^{W}\big(\bar{w}_t+\alpha_t(h(\bar{w}_t)+
M_{t+1}+\mathnormal{l}_t)-\bar{w}_t\big)}{\alpha_t} \\
&= 
\bar{w}_t+\alpha_t\Big(\hat{\Gamma}_{\bar{w}_t}^{W}(h(\bar{w}_t))+
\hat{\Gamma}_{\bar{w}_t}^{W}(M_{t+1}) \\
&+ \hat{\Gamma}_{\bar{w}_t}^{W}(\mathnormal{l}_t)+o(\alpha_t) \Big),
\end{split}
\end{equation}
where $\Gamma^{W}$ and $\hat{\Gamma}_{\bar{w}_t}^{W}$ are defined in
previous section.
 
Then, a few observations are in order: 
($i$) $\hat{\Gamma}_{\bar{w}_t}^{W}(h(\bar{w}_t))$is a Lipschitz
continuous function in $\bar{w}_t$, which follows from the hypothesis of the Theorem.
($ii$) $\hat{\Gamma}_{\bar{w}_t}^{W}(M_{t+1})$ is a truncated martingale
difference noise. Indeed, it is easy to verify that the noise sequence
$\{M_{t+1}\}_{t\in\mathbb{N}}$ is a martingale-difference sequence with respect
to the filtration $\{\mathcal{F}_t\}_{t\in{N}}$, i.e., $\forall t\in\mathbb{N}$,
$M_{t+1}$ is $\mathcal{F}_{t+1}-$measurable and integrable, and 
$\mathbb{E}[M_{t+1}|\mathcal{F}_{t}]=0$ a.s., $\forall t\in\mathbb{N}$. Also,
since $\hat{\Gamma}^{W}_{\bar{w}_t}$ is a bounded linear operator, we have
$\hat{\Gamma}^{W}_{\bar{w}_t}(M_{t+1})$ to be
$\mathcal{F}_{t+1}-$measurable and integrable, $\forall t\in\mathbb{N}$. 
Further, $\exists K_0\in(0,\infty)$, such that
\begin{equation*}
\mathbb{E}\Big[||\hat{\Gamma}_{\bar{w}_t}^{W}(M_{t+1})||^2|\mathcal{F}_t
\Big]\leq K_0(1+||\bar{w}_t||^2)\quad a.s.,
\end{equation*}
which follows directly from the finiteness of the Markov chain
\cite{levin2017markov} and the boundary $\partial W$ is smooth.
($iii$) For the bias, we have
\begin{equation*}
\begin{split}
\big\Vert\hat{\Gamma}_{\bar{w}_t}^{W}(\mathnormal{l}_t)\big\Vert
&= \Big\Vert\lim_{\epsilon\to
0}\frac{\Gamma^{W}(\bar{w}_t+\epsilon\mathnormal{l}_t)-\bar{w}_t}{\epsilon}\Big\Vert
\\
&\leq \lim_{\epsilon\to 0}\frac{\big\Vert
\Gamma^{W}(\bar{w}_t+\epsilon\mathnormal{l}_t)-\Gamma^{W}(\bar{w}_t)
\big\Vert}{\epsilon} \\
&\leq \lim_{\epsilon\to 0}\frac{\big\Vert
\bar{w}_t+\epsilon\mathnormal{l}_t-\bar{w}_t
\big\Vert}{\epsilon} = \Vert\mathnormal{l}_t\Vert , \\
\end{split}
\end{equation*}
and $\hat{\Gamma}^{W}_{\bar{w}_t}(\mathnormal{l}_{t})\to 0$ as
$t\to\infty$ a.s., which follows directly from the ergodicity and finiteness of
the underlying Markov chain \cite{levin2017markov}.
($iv$) $o(\alpha_t)\to 0$ as $t\to \infty$.
($v$) The iterates of $\bar{w}_t$ remain bounded a.s., i.e., 
\begin{equation*}
\sup_{t\in\mathbb{N}}\Vert \bar{w}_t \Vert<\infty, a.s.,
\end{equation*}
since $\bar{w}_t \in W, \forall t\in\mathbb{N}$ and $W$ is compact.

Thus, by appealing to Theorem $2$, Chapter $2$ of Borkar \cite{borkar2009stochastic}, we conclude
that the stochastic recursion (\ref{w0}) converges to the asymptotically stable equilibria of the
ODE contrained inside $W$, which is 
\begin{equation*}
\frac{d}{dt}\bar{w}(t)=\hat{\Gamma}^{W}_{\bar{w}(t)}(h(\bar{w}_t))
=\hat{\Gamma}^{W}_{\bar{w}(t)}(-\frac{1}{2}\nabla_{\bar{w}}
\mathcal{L}_{slow})(\bar{w}(t)),
\end{equation*}
where $\bar{w}(0)\in\mathring{W}$ and $t\in\mathbb{R}_+$.

\textbf{In the fast optimization process}, the fast optimization is
a traditional TD algorithm, our point of interest for the on-policy algorithm.
Stability analysis of on-policy TD has been given by Sutton
\cite{sutton1998reinforcement}, which converges to a TD fixed point. Further, we
obtain a fixed point given a stable $\bar{w}_t\equiv\bar{w}^*$, which follows
from the TD fixed point of the fast optimization
\begin{equation*} 
\Phi_{\bar{w}^*,D}\theta^*=\Pi_{\bar{w}^*,D}\mathbb{T}(\Phi_{\bar{w}^*,D}\theta^*),
\end{equation*}  
which is  the final fixed point of two-timescale optimization.
\end{IEEEproof}

We prove that the online dictionary construction is convergent, and the
two-timescale optimization is
convergent under the assumption that the dictionary is fixed.  Our analysis is
not enough to guarantee that the OAKTD algorithm is convergent.  However, this
analysis is acceptable, because under the assumption that the  dictionary
construction converges first, i.e., after the dictionary is fixed, the
convergence of OAKTD is guaranteed. We will verify this  assumption in the
next section.

%% file: experiments.tex
\section{Experiments}
In this section, we conduct experiments on four classic control tasks with
continuous state spaces: Mountain Car, Acrobot, CartPole and Puddle World
\cite{sutton1996generalization}, \cite{sutton1998reinforcement}, which are all
public benchmarks for studying online reinforcement learning algorithms. To
accurately evaluate performance of the attentive kernel-based model as sparse
representation, we compared our proposed OAKTD \footnote{Note that OAKTD is a
learning algorithm for prediction. In this paper, we use its on-policy version
to learn for control, and adopt $\epsilon-$greedy policy. But we still use the
abbreviation OAKTD rather than OAK-SARSA.} with OKTD, OSKTD and TD with
Tile Coding. In addition, to reflect characteristics of the attention mechanism as a
degree of sparsification, we representatively visualize the attention on the
kernel in Mountain Car since it is a $2$-dimensional task similar to Puddle
World.

\begin{figure*}[ht]
\centering
	\subfigure{
	\begin{minipage}[t]{0.45\linewidth}
	\centering
	\includegraphics[scale=0.38]{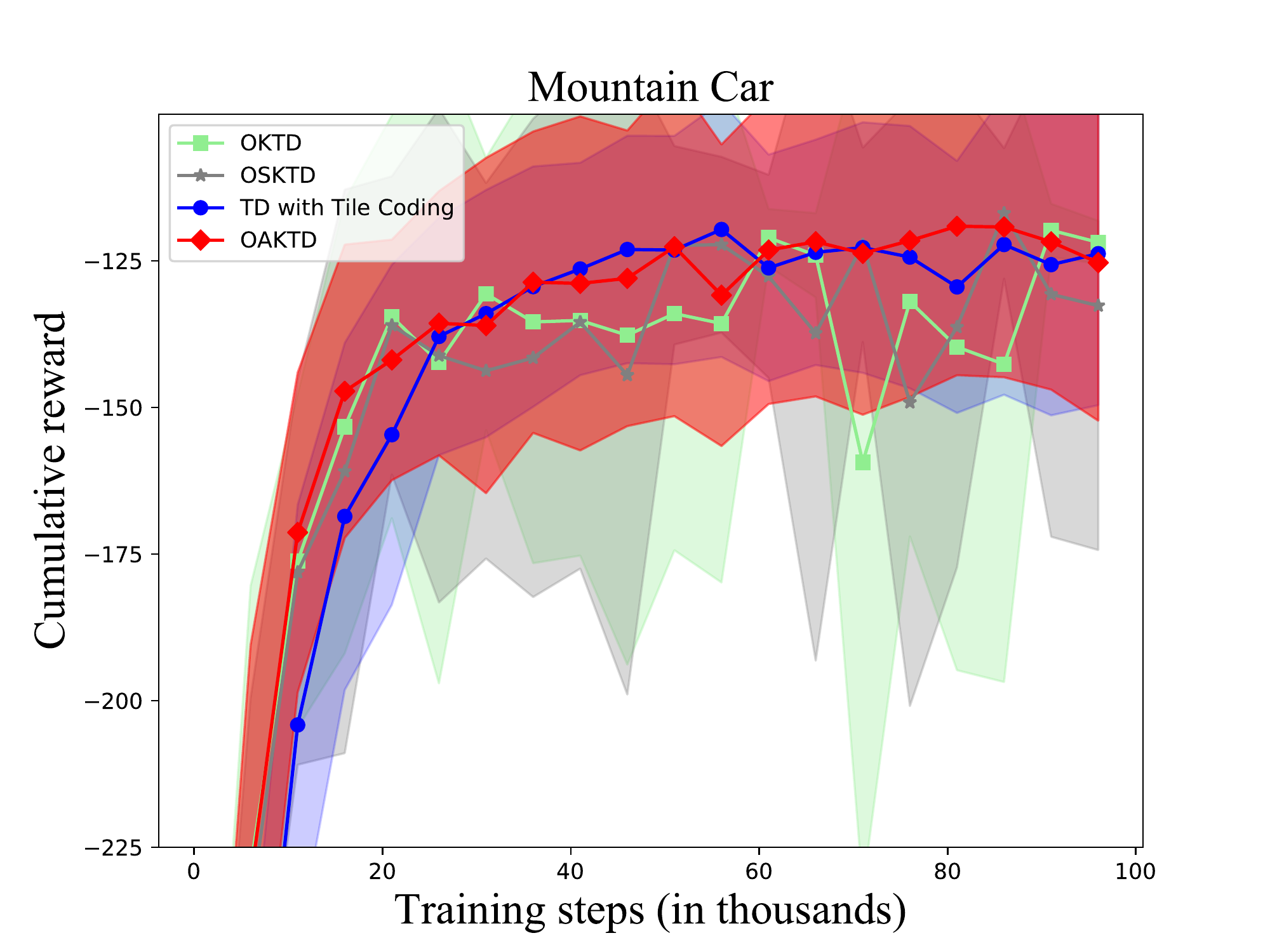}
	\end{minipage}
	}
	\subfigure{
	\begin{minipage}[t]{0.45\linewidth}
	\centering
	\includegraphics[scale=0.38]{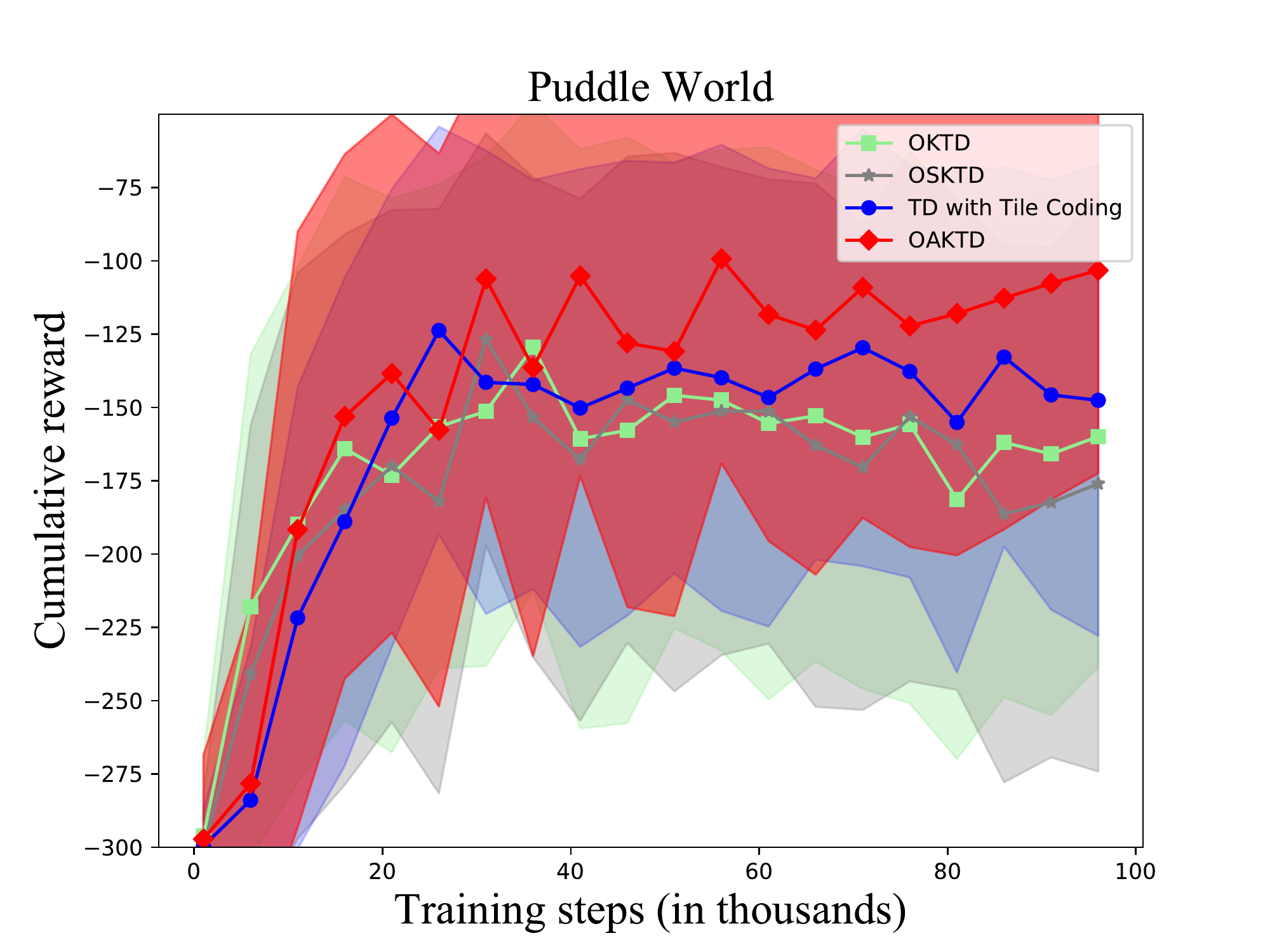}
	\end{minipage}
	}
	
	\subfigure{
	\begin{minipage}[t]{0.45\linewidth}
	\centering
	\includegraphics[scale=0.38]{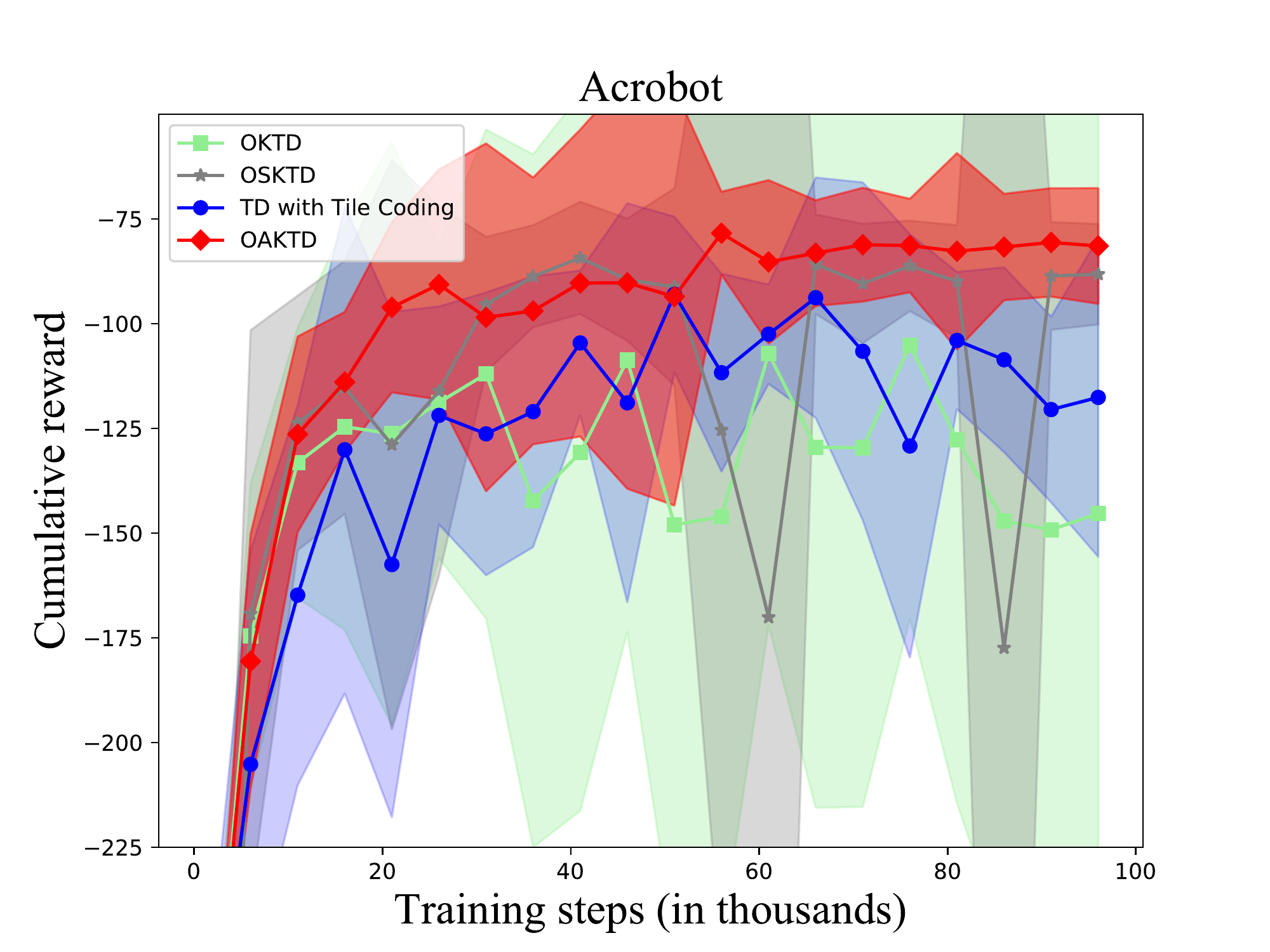}
	\end{minipage}
	}
	\subfigure{
	\begin{minipage}[t]{0.45\linewidth}
	\centering
	\includegraphics[scale=0.38]{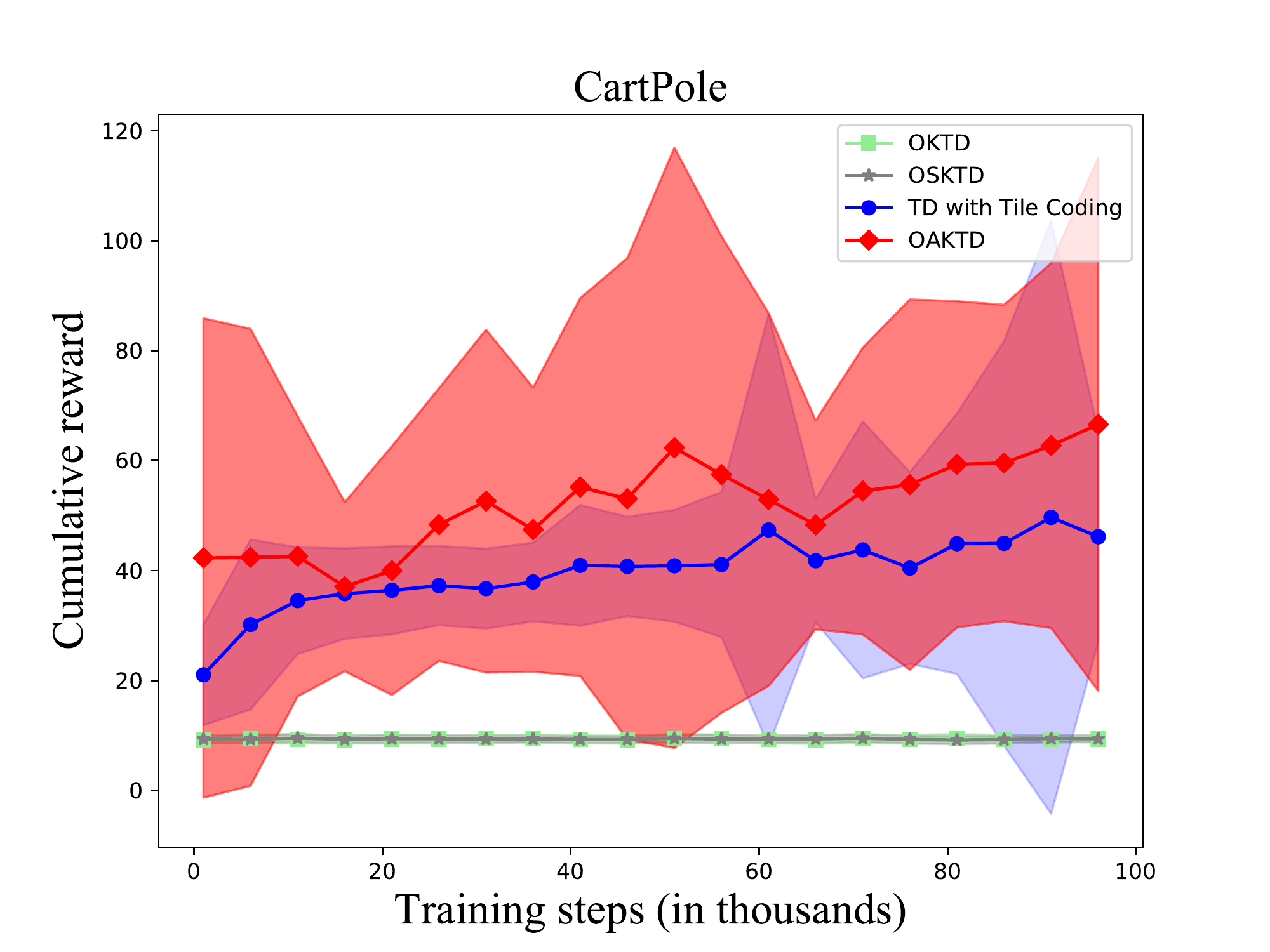}
	\end{minipage}
	} 
\caption{Learning curves in Mountain Car, Acrobot, CartPole and Puddle
World. The abscissa is the number of training steps, and the ordinate is the
cumulative reward.}
\label{rewards}
\end{figure*}

\begin{figure*}[ht]
\centering
	\subfigure{
	\begin{minipage}[t]{0.45\linewidth}
	\centering
	\includegraphics[scale=0.40]{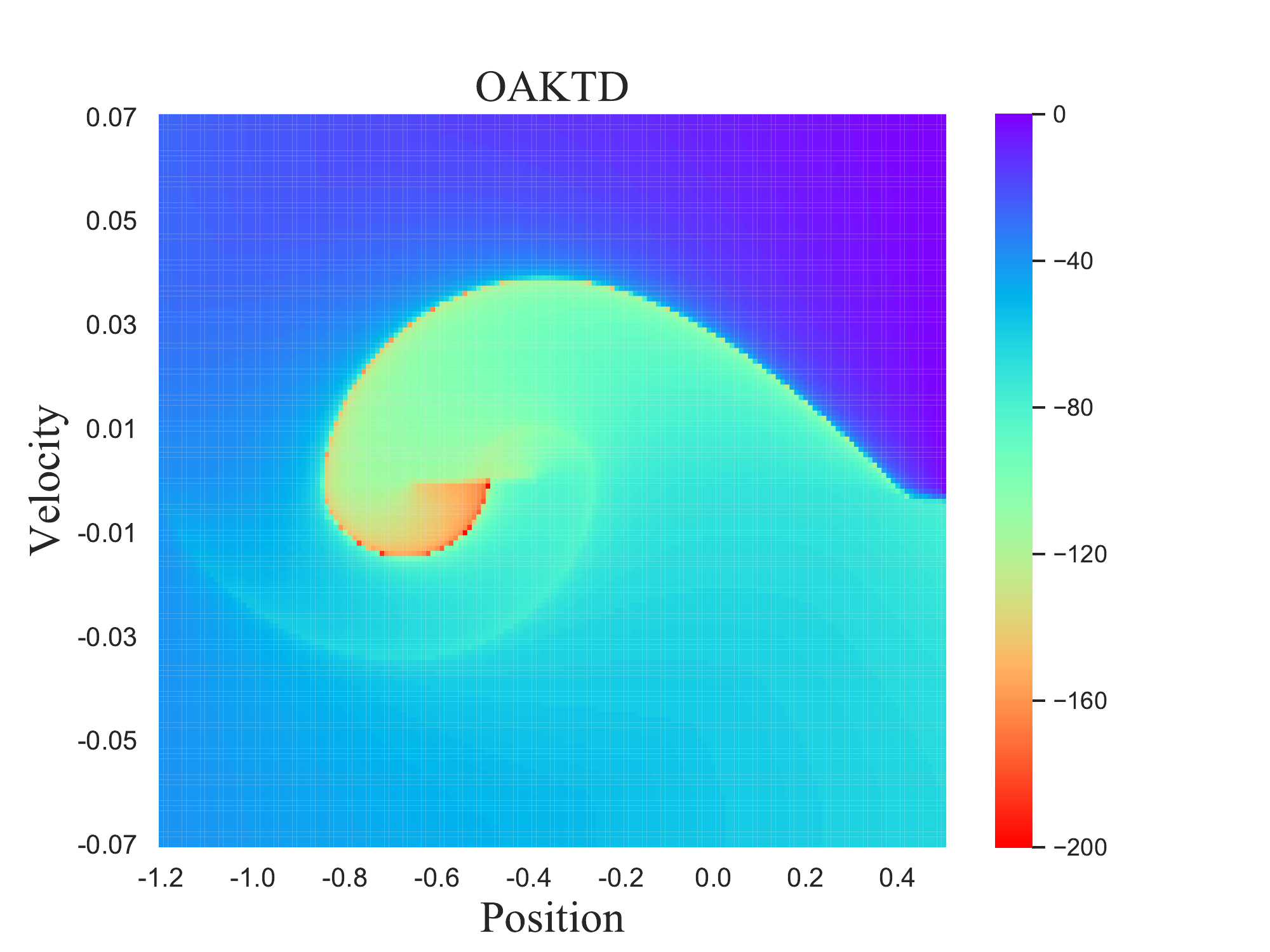}
	\end{minipage}
	}
\centering
	\subfigure{%
	\begin{minipage}[t]{0.45\linewidth}
	\centering
	\includegraphics[scale=0.40]{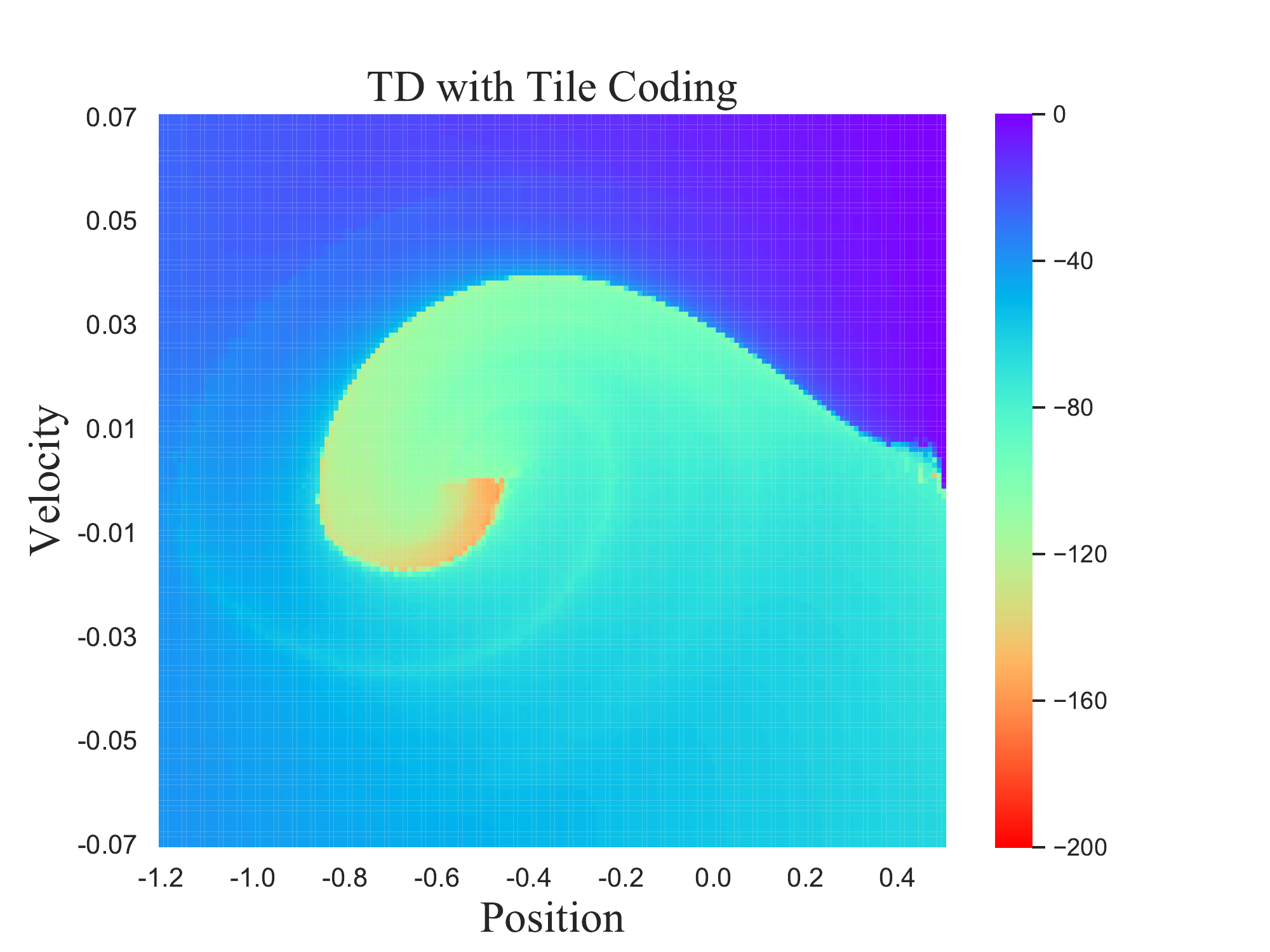}
	\end{minipage}
	}
\caption{Visualized in the color images are rewards on the whole state
space. The left shows the performance of OAKTD and the right shows the
performance of TD with Tile Coding.}
\label{rewards_all}
\end{figure*}

\subsection{Descriptions of Control Tasks}
\textbf{The Mountain Car} is the problem about how to drive an car up to the
top of a hill. The difficulty is that gravity is much stronger than the car's engine and thus
the car cannot accelerate directly to up the hill. There are three alternative
actions: full throttle forward ($+1$), full throttle reverse ($-1$), and zero
throttle ($0$). Its position $x_{t+1}$ can be updated by it last position
$x_{t}$ and velocity $\dot{x_t}$, which is, $$x_{t+1}=\text{bound}[x_t+\dot{x}_{t+1}],$$
where $\dot{x}_{t+1}=\text{bound}[\dot{x}_t+0.001a_t -0.0025 \cos(3x_t)]$,
the bound operation enforces $-1.2 \leq x_{t+1} \leq 0.5$ and $-0.07 \leq
\dot{x}_{t+1} \leq 0.07$. In addition, when $x_{t+1}$ reaches the left bound,
$\dot{x}_{t+1}$ will be reset to zero. The reward is $-1$ for all states except the
goal state at the top of the hill in which the episode ends with a reward $0$. The
discount factor is set to $0.99$, and each episode starts from a random position
$x_0\in[-0.6,-0.4)$ with zero velocity.

\textbf{The Acrobot} is a two-link under-actuated robot. Our goal is to swing
the end-effector at a height at least the length of one link above the base.
There are also three discrete actions: apply positive torque ($1$), apply
negative torque ($-1$), and apply no torque ($0$). The state consists of the
two rotational joint angles and their velocities. The reward is $-1$ for all
the states except the goal state in which the episode ends with a reward $0$.
The discount factor is set to $0.99$. The agent is initialized in a downward
vertical position.

\textbf{The Cartpole} is a the classic inverted pendulum problem with a center
of gravity above its pivot point. It is unstable and can be controlled by moving
the pivot point under the center of mass. The state of Cartpole consists of its
position, velocity, angle, and angular velocity. The cart only has two possible
actions: move to the left or move to the right. A reward of $+1$ is provided for
every timestep that the pole remains upright. The discount factor is set to
$1$. The goal is to keep the cartpole balanced by applying appropriate forces
to a pivot point.
 
\textbf{The Puddle World} is a task to seek for a goal position in a limited
scene. The environment is a $1\times 1$ square, where the puddles are $0.1$ in
radius and are located at center points $(0.1, 0.75)$ to $(0.45, 0.75)$ and $(0.45,
0.4)$ to $(0.45, 0.8)$. The agent starts at position $(0.2, 0.4)$ and can choose
five actions, up, down, left, right and stop, which moved approximately $0.05$
in these directions under the bounder $[0, 1]$. In addition, a random gaussian
noise with standard deviation $0.01$ was also added to the motion along both
dimensions. The reward is $-1$ for each time step plus additional penalties if
the agent does not enter the puddles. These penalties were $-400$ times the
distance to the nearest edge of the puddles regions. The discount factor is set
to $0.999$.

\subsection{Experimental Settings}
\textbf{In the Mountain Car task}, for fair comparison as possible, we adopt
$\epsilon$-greedy exploration in the learning process, and set the greedy parameter $\epsilon$ to
$\epsilon_{t}=0.9999^t$, i.e., it will gradually decreases to zero as those
methods proceed. The learning rate of linear approximation is uniformly set to
$0.01\times 0.999999^t$, which is also the fast learning rate in OAKTD. In
online kernel-based methods, we use a Gaussian kernel with parameter $\sigma_2=1$. It is
necessary to standardize the states for eliminating the effect of feature scale, and the
$k$-th element can be defined as
$\frac{\chi_k(s)-\min(\chi_k(s))}{\max(\chi_k(s))-\min(\chi_k(s))}$.
Then, the parameters of online dictionary construction are set to $\mu_1=0.1$,
$\sigma_1=1$. For the specific parameters of different methods, we set to the
slow learning rate $\alpha_t=0.05\times 0.99999^t$, the fast learning rate
$\beta_t=0.01\times 0.999999^t$. In OSKTD, we set to selective function
$\mu_2=1$. In addition, in TD with Tile Coding. we set the number of tilings to
$16$, the size of every tiling to $256$.

\textbf{In the Acrobot task}, the learning rate is uniformly set to
$0.001\times 0.999999^t$. For online kernel-based methods, the
parameters of dictionary construct are set to $\mu_1=0.3, \sigma_1=1$ and the
parameter of kernel function is set to $\sigma_2=1$. For the specific parameters
of different methods, we set to the slow learning rate $\alpha_t=0.1\times
0.99999^t$, the fast learning rate $\beta_t=0.001\times 0.999999^t$ in OAKTD. In
OSKTD, we set selective function to $\mu_2=1$. In addition, in TD with Tile
Coding. we set the number of tilings to $16$, the size of every tiling to
$4096$.

\textbf{In the CartPole task}, the learning rate is uniformly set to
$0.01\times 0.9999999^t$. The parameters of dictionary construct are
same as the setting of the Acrobot. For the specific parameters of different
methods, in OAKTD, we set to the slow learning rate $\alpha_t=0.1\times
0.999999^t$ and the fast learning rate $\beta_t=0.01\times 0.9999999^t$. In
OSKTD, we set to selective function $\mu_2=1$. In addition, in TD with Tile
Coding, we set the number of tilings to $16$, the size of every tiling to
$4096$.

\textbf{In the Puddle World task}, the learning rate is uniformly set to
$0.01\times 0.999999^t$. the parameter of dictionary construct is set
to $\mu_1=0.1, \sigma_1=1$ and the parameter of kernel function is set to
$\sigma_2=0.1$.
In OAKTD, we set to the slow learning rate $\alpha_t=0.1\times 0.99999^t$ and the
fast learning rate $\beta_t=0.01\times 0.999999^t$. In OSKTD, we set to
selective function $\mu_2=1.4$. In addition, in TD with Tile Coding, we set the
number of tilings to $16$, the size of every tiling to $512$.

\subsection{Results and Analysis}
In the learning process, each algorithm runs over $50$ times, each 
with one million steps. For online dictionary construction in our
proposed OAKTD algorithm, we record 
the mean and std of the dictionary size after convergence,
and the mean and std of the time steps (in percentage of learning steps) taken
when the dictionary converges, as shown in Table \ref{dictionary}.
It is easy to find that the convergence steps of the online dictionary
construction are far fewer than the learning steps.
It verifies the assumption that the  online dictionary
construction converges first. 

\begin{table}
\centering
\caption{Dictionary stability of OAKTD in various environments.}
\label{dictionary}
\begin{tabular}{ccc}
\hline
\multirow{2}{*}{Environment} & \multirow{2}{*}{Dictionary Size} &
\multirow{2}{*}{$\frac{\text{Convergence Steps}}{\text{Training
Steps}}\times 100\%$}
\\
\\
\hline
Mountain Car  & $80.35 \pm 4.32$ & $0.59\pm 0.22$ 
\\
Acrobot       & $5.20   \pm 2.54$ & $0.23\pm 0.18$ 
\\
Cartpole      & $90.85 \pm 18.23$ & $0.29\pm 0.10$
\\
Puddle World  & $181.65 \pm 7.13$ & $0.36\pm 0.13$
\\
\hline
\end{tabular}
\end{table}

In addition, in order to effectively evaluate the performance of different
algorithms, we test the performance of the algorithm every thousand learning steps. The
learning curves of several algorithms for different control tasks are shown 
for the first $100,000$ steps in
Fig. \ref{rewards}. In Mountain Car, from the perspective of cumulative rewards,
compared with OKTD and OSKTD, OAKTD has higher mean and lower standard deviation. Compared
with TD with Tile Coding, OAKTD has a faster convergence rate and
approximate mean, but their final convergence results are very similar. To
further reflect the behavior of OAKTD, we sample the total state space that is a
set of $171\times 141$ points evenly spaced in the position-velocity space, as
the initial state of Mountain Car. In particular, the position (velocity)
interval is evenly divided into subintervals of length $0.01$ ($0.001$). As
shown in Fig. \ref{rewards_all}, TD with Tile Coding has more light streaks in
the cool colors, which shows agent needs more steps in these areas. OAKTD has
more smooth performance.
Moveover, the number of OAKTD parameters is also far less than TD with Tile
Coding. After the above analysis,  OAKTD has the best
performance for Mountain Car.

\begin{figure}[ht]
\centering
\includegraphics[scale=0.35]{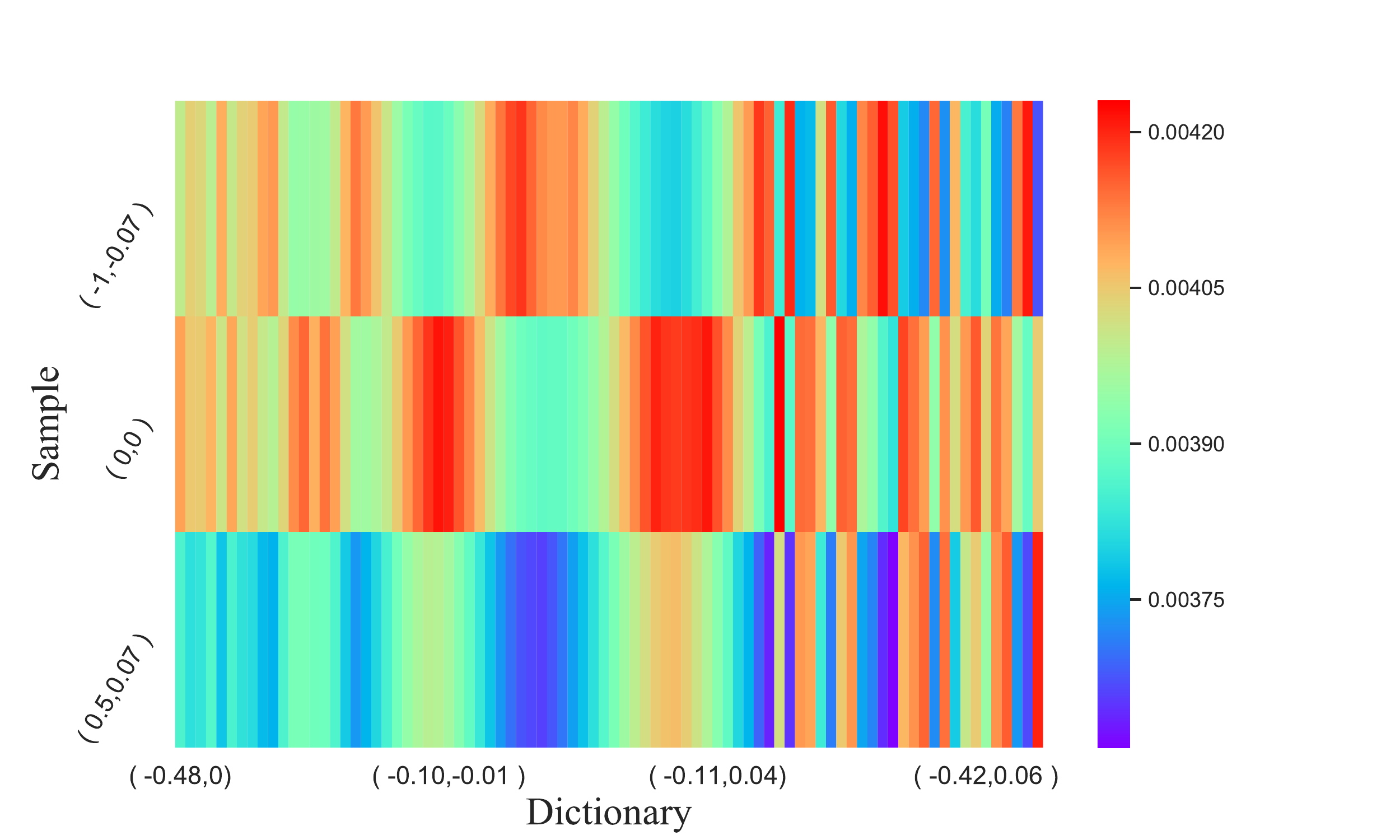}
\caption{Visualized in the color image is the state's attention on the kernels.}
\label{attention}
\end{figure}

\begin{figure}[ht]
\centering
\includegraphics[scale=0.41]{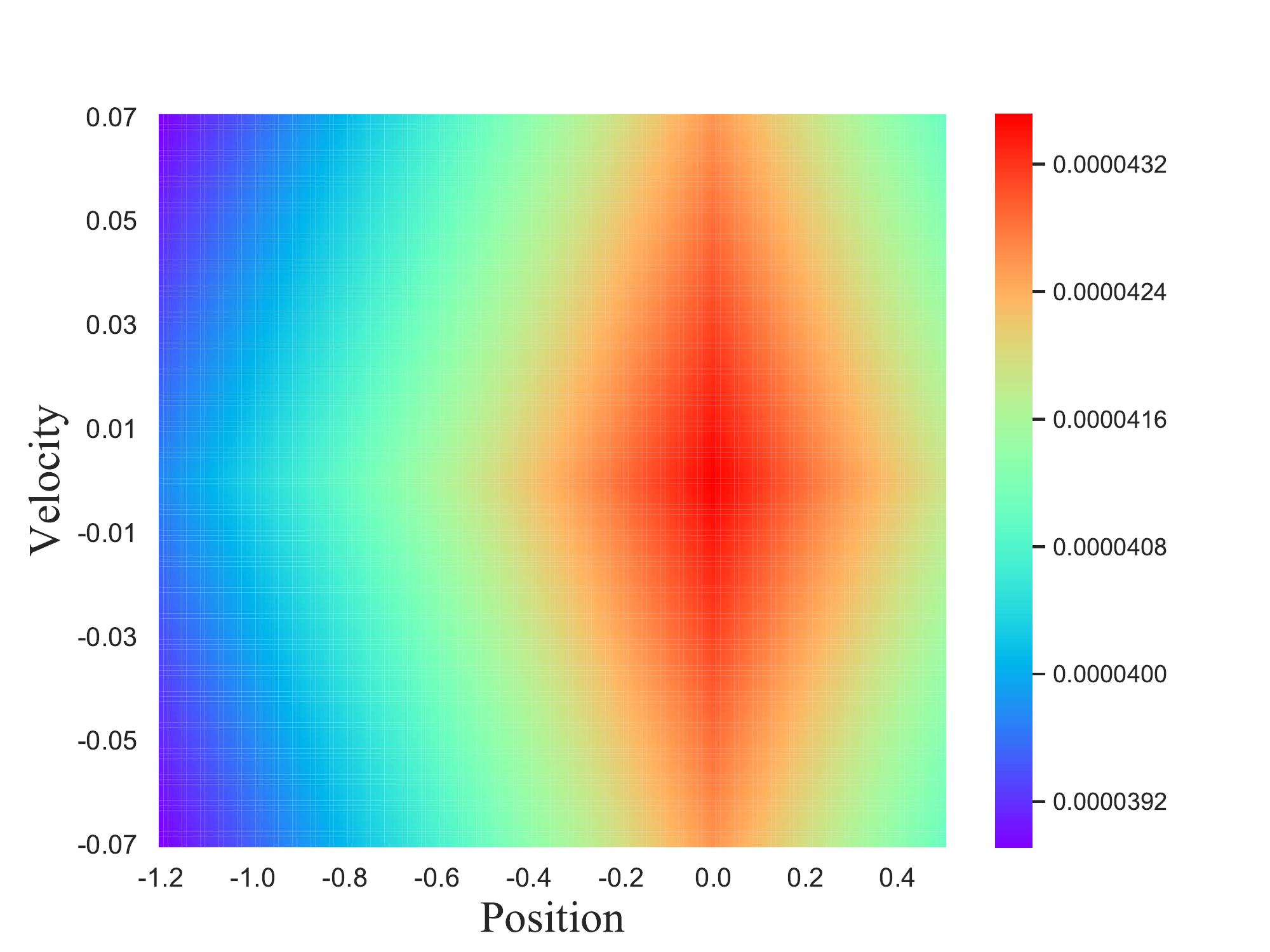}
\caption{Visualized in the color image is the state's attention on the whole
state space.}
\label{individual_attentions}
\end{figure}

In the other tasks, the performance of TD with Tile Coding is
relatively declined in $2$-dimensional Puddle World, and $4$-dimensional
Acrobot and CartPole, compared to $2$-dimensional Mountain Car. OKTD and
OSKTD have the worst standard deviation in Acrobot and Puddle World, and
don't even work in CartPole. However, OAKTD has best mean and standard
deviation. The steady performance of OAKTD in the three control problems further
expands the advantage of attentive function in online RL. Furthermore, 

To accurately describe the attention on the kernels, based on behavior of OAKTD
in Mountain Car, we sample three states $(-1, -0.07),(0, 0),(0.5, 0.07)$, a
dictionary, and attentive parameters $w=(-0.13, -0.04)$.
After standardization, visualization of attention is shown in Fig. \ref{attention}. It
vividly shows that attention is distributed in the kernel functions as the
degree of sparsification. To further describe the distribution situation, we
selected state $(0, 0)$ and visualized its attention distribution in the whole
state space. As shown in Fig. \ref{individual_attentions}, we can observe that
attention is focused around itself and spreads outward. The experimental results
show the attentive function can be well suited for sparse representation
learning, and attentive kernel-based VFA can be more suitable for online RL.

%% file: conclusion.tex
\section{Conclusion and Future Work}
In this paper, we construct an attentive kernel-based model and proposed a
stable Online Attentive Kernel-based Temporal Difference (OAKTD) learning
algorithm based on two-timescale optimization in order to simplify the model
and alleviate catastrophic interference for online Reinforcement Learning (RL).
Furthermore, we prove the convergence of our proposed algorithm. For the four
classic control tasks (Mountain Car, Acrobot, CartPole and Puddle World), our
experimental results verify that compared with OSKTD, OKTD and TD with Tile
Coding, OAKTD performed the best.

For future study, we will focus on the following aspects: ($i$) to extend our
OAKTD algorithm to handle continuous action RL problems, e.g., combining policy
gradient or its variants \cite{lillicrap2015continuous, mnih2016asynchronous};
($ii$) other ways of sparse representation that satisfy the proposed four
characteristics: learnable, non-prior, non-truncated and explicit; ($iii$) to
guarantee the stability of the algorithm,  we utilize
two-timescale optimization and we wish to know if other optimizations (e.g.,
semi-gradient method) could result in a convergence guarantee? ($iv$) since our
proposed attentive kernel-based model is applied to the temporal difference
learning algorithm, it could obviously be applied to other machine learning
algorithms as well, e.g., attentive Support Vector Machine (SVM), attentive
Support Vector Regression (SVR) etc.

\section*{Acknowledgement}
The authors would like to thank the anonymous referees
and the editor for their helpful comments and suggestions.

%% file: document.bbl
\begin{thebibliography}{10}
\providecommand{\url}[1]{#1}
\csname url@samestyle\endcsname
\providecommand{\newblock}{\relax}
\providecommand{\bibinfo}[2]{#2}
\providecommand{\BIBentrySTDinterwordspacing}{\spaceskip=0pt\relax}
\providecommand{\BIBentryALTinterwordstretchfactor}{4}
\providecommand{\BIBentryALTinterwordspacing}{\spaceskip=\fontdimen2\font plus
\BIBentryALTinterwordstretchfactor\fontdimen3\font minus
  \fontdimen4\font\relax}
\providecommand{\BIBforeignlanguage}[2]{{%
\expandafter\ifx\csname l@#1\endcsname\relax
\typeout{** WARNING: IEEEtran.bst: No hyphenation pattern has been}%
\typeout{** loaded for the language `#1'. Using the pattern for}%
\typeout{** the default language instead.}%
\else
\language=\csname l@#1\endcsname
\fi
#2}}
\providecommand{\BIBdecl}{\relax}
\BIBdecl

\bibitem{sutton1998reinforcement}
R.~S. Sutton and A.~G. Barto, \emph{Reinforcement learning: An
  introduction}.\hskip 1em plus 0.5em minus 0.4em\relax Cambridge, MA: USA: MIT
  Press, 2018.

\bibitem{liu2019sparse}
V.~Liu, ``Sparse representation neural networks for online reinforcement
  learning,'' Ph.D. dissertation, University of Alberta, 2019.

\bibitem{liu2019utility}
V.~Liu, R.~Kumaraswamy, L.~Le \emph{et~al.}, ``The utility of sparse
  representations for control in reinforcement learning,'' in \emph{Proc. 33rd
  AAAI Conf. Artif. Intell.}, Honolulu, Hawaii, USA, January 2019, pp.
  4384--4391.

\bibitem{sutton1988learning}
R.~S. Sutton, ``Learning to predict by the methods of temporal differences,''
  \emph{Mach. Learn.}, vol.~3, no.~1, pp. 9--44, 1988.

\bibitem{lecun2015deep}
Y.~LeCun, Y.~Bengio, and G.~Hinton, ``Deep learning,'' \emph{Nature}, vol. 521,
  no. 7553, pp. 436--44, May. 2015.

\bibitem{sutton2009fast}
R.~S. Sutton, H.~R. Maei, D.~Precup \emph{et~al.}, ``Fast gradient-descent
  methods for temporal-difference learning with linear function
  approximation,'' in \emph{Proc. 26th Int. Conf. Mach. Learn.}, Montreal,
  Quebec, Canada, Jun. 2009, pp. 993--1000.

\bibitem{sun2020finite}
J.~Sun, G.~Wang, G.~B. Giannakis \emph{et~al.}, ``Finite-time analysis of
  decentralized temporal-difference learning with linear function
  approximation,'' in \emph{Proc. 23rd Int. Conf. Artif. Intell. Stat.},
  Palermo, Sicily, Italy, Aug. 2020, pp. 4485--4495.

\bibitem{lee2019target}
D.~Lee and N.~He, ``Target-based temporal-difference learning,'' in \emph{Proc.
  36th Int. Conf. Mach. Learn.}, Long Beach, California, USA, Jun. 2019, pp.
  3713--3722.

\bibitem{ormoneit2002kernel}
D.~Ormoneit and {\'S}.~Sen, ``Kernel-based reinforcement learning,''
  \emph{Mach. Learn.}, vol.~49, no. 2-3, pp. 161--178, 2002.

\bibitem{chen2013online}
X.~Chen, Y.~Gao, and R.~Wang, ``Online selective kernel-based temporal
  difference learning,'' \emph{IEEE Trans. Neural Netw. Learn. Syst.}, vol.~24,
  no.~12, pp. 1944--1956, 2013.

\bibitem{sutton1996generalization}
R.~S. Sutton, ``Generalization in reinforcement learning: Successful examples
  using sparse coarse coding,'' in \emph{Proc. Adv. Neural Inf. Process. Syst.
  9}, Denver, CO, USA, Dec. 1996.

\bibitem{krawiec2011learning}
K.~Krawiec and M.~G. Szubert, ``Learning n-tuple networks for othello by
  coevolutionary gradient search,'' in \emph{Proc. 13th Ann. Conf. Genet.
  Evolut. Comput.}, Dublin, Ireland, Jul. 2011, pp. 355--362.

\bibitem{nair2010rectified}
V.~Nair and G.~E. Hinton, ``Rectified linear units improve restricted boltzmann
  machines,'' in \emph{Proc. 27th Int. Conf. Mach. Learn.}, Haifa, Israel, Jun.
  2010, pp. 807--814.

\bibitem{srivastava2014dropout}
N.~Srivastava, G.~Hinton, A.~Krizhevsky \emph{et~al.}, ``Dropout: a simple way
  to prevent neural networks from overfitting,'' \emph{J. Mach. Learn. Res.},
  vol.~15, no.~1, pp. 1929--1958, 2014.

\bibitem{makhzani2013k}
A.~Makhzani and B.~Frey, ``K-sparse autoencoders,'' in \emph{Proc. 2nd Int.
  Conf. Learn. Repr.}, Banff, AB, Canada, Apr. 2014.

\bibitem{makhzani2015winner}
A.~Makhzani and B.~J. Frey, ``Winner-take-all autoencoders,'' in \emph{Proc.
  Adv. Neural Inf. Process. Syst. 28}, Montreal, Quebec, Canada, December 2015,
  pp. 2791--2799.

\bibitem{park2007l1}
M.~Y. Park and T.~Hastie, ``L1-regularization path algorithm for generalized
  linear models,'' \emph{J. Royal Stat. Soc.}, vol.~69, no.~4, pp. 659--677,
  2007.

\bibitem{girosi1995regularization}
F.~Girosi, M.~Jones, and T.~Poggio, ``Regularization theory and neural networks
  architectures,'' \emph{Neural Comput.}, vol.~7, no.~2, pp. 219--269, 1995.

\bibitem{xu2015show}
K.~Xu, J.~Ba, R.~Kiros \emph{et~al.}, ``Show, attend and tell: Neural image
  caption generation with visual attention,'' in \emph{Proc. 32nd Int. Conf.
  Mach. Learn.}, Lille, France, Jul. 2015, pp. 2048--2057.

\bibitem{mnih2014recurrent}
V.~Mnih, N.~Heess, A.~Graves \emph{et~al.}, ``Recurrent models of visual
  attention,'' in \emph{Proc. Adv. Neural Inf. Process. Syst. 27}, Montreal,
  Quebec, Canada, Dec. 2014, pp. 2204--2212.

\bibitem{bahdanau2015neural}
D.~Bahdanau, K.~Cho, and Y.~Bengio, ``Neural machine translation by jointly
  learning to align and translate,'' in \emph{Proc. 3rd Int. Conf. Learn.
  Repr.}, San Diego, CA, USA, May. 2014.

\bibitem{chung2018two}
W.~Chung, S.~Nath, A.~Joseph \emph{et~al.}, ``Two-timescale networks for
  nonlinear value function approximation,'' in \emph{Proc. 7th Int. Conf.
  Learn. Repr.}, New Orleans, LA, USA, May. 2019.

\bibitem{scholkopf2001generalized}
B.~Sch$\ddot{o}$lkopf, R.~Herbrich, and A.~J. Smola, ``A generalized
  representer theorem,'' in \emph{Proc. Int. Conf. Comput. Learn. Theory},
  Berlin, Heidelberg, 2001, pp. 416--426.

\bibitem{liu2009information}
W.~Liu, I.~Park, and J.~C. Principe, ``An information theoretic approach of
  designing sparse kernel adaptive filters,'' \emph{IEEE Trans. Neural Netw.},
  vol.~20, no.~12, pp. 1950--1961, 2009.

\bibitem{baird1995residual}
L.~Baird, ``Residual algorithms: Reinforcement learning with function
  approximation,'' in \emph{Proc. 12th Int. Conf. Mach. Learn.}, Tahoe City,
  California, USA, Jul. 1995, pp. 30--37.

\bibitem{Hamid2009convergent}
H.~R. Maei, C.~Szepesv{\'{a}}ri, S.~Bhatnagar \emph{et~al.}, ``Convergent
  temporal-difference learning with arbitrary smooth function approximation,''
  in \emph{Proc. Adv. Neural Inf. Process. Syst. 22}, Vancouver, British
  Columbia, Canada, Dec. 2009, pp. 1204--1212.

\bibitem{engel2004kernel}
Y.~Engel, S.~Mannor, and R.~Meir, ``The kernel recursive least-squares
  algorithm,'' \emph{IEEE Trans. Neural Netw. Learn. Syst.}, vol.~52, no.~8,
  pp. 2275--2285, 2004.

\bibitem{xu2007kernel}
X.~Xu, D.~Hu, and X.~Lu, ``Kernel-based least squares policy iteration for
  reinforcement learning,'' \emph{IEEE Trans. Neural Netw. Learn. Syst.},
  vol.~18, no.~4, pp. 973--992, 2007.

\bibitem{borkar1997stochastic}
V.~S. Borkar, ``Stochastic approximation with two time scales,'' \emph{Syst. \&
  Cont. Lett.}, vol.~29, no.~5, pp. 291--294, 1997.

\bibitem{teschl2012ordinary}
G.~Teschl, \emph{Ordinary Differential Equations and Dynamical Systems}.\hskip
  1em plus 0.5em minus 0.4em\relax American Mathematical Society, 2012.

\bibitem{kushner2012stochastic}
H.~J. Kushner and D.~S. Clark, \emph{Stochastic approximation methods for
  constrained and unconstrained systems}.\hskip 1em plus 0.5em minus
  0.4em\relax Springer Science \& Business Media, 2012.

\bibitem{borkar2009stochastic}
V.~S. Borkar, \emph{Stochastic approximation: a dynamical systems
  viewpoint}.\hskip 1em plus 0.5em minus 0.4em\relax Springer, 2009.

\bibitem{levin2017markov}
D.~A. Levin and Y.~Peres, \emph{Markov chains and mixing times}.\hskip 1em plus
  0.5em minus 0.4em\relax American Mathematical Society, 2017.

\bibitem{lillicrap2015continuous}
T.~P. Lillicrap, J.~J. Hunt, A.~Pritzel \emph{et~al.}, ``Continuous control
  with deep reinforcement learning,'' in \emph{Proc. 4th Int. Conf. Learn.
  Repr.}, San Juan, Puerto Rico, May. 2016.

\bibitem{mnih2016asynchronous}
V.~Mnih, A.~P. Badia, M.~Mirza \emph{et~al.}, ``Asynchronous methods for deep
  reinforcement learning,'' in \emph{Proc. 33rd Int. Conf. Mach. Learn.}, New
  York City, NY, USA, Jun. 2016, pp. 1928--1937.

\end{thebibliography}
